\documentclass[11pt]{article}

\usepackage[final]{acl}

\usepackage{times}
\usepackage{latexsym}
\usepackage[T1]{fontenc}
\usepackage[utf8]{inputenc}
\usepackage{microtype}
\usepackage{inconsolata}

\usepackage{graphicx}
\usepackage{booktabs}
\usepackage{array}
\usepackage{amsmath, amssymb}
\usepackage{xcolor}
\usepackage{enumitem}
\usepackage{placeins}
\usepackage{url}
\usepackage{verbatim}
\usepackage{pifont}
\usepackage{fontawesome5}
\usepackage{multirow}

\newcommand{\cmark}{\ding{51}}
\newcommand{\xmark}{\ding{55}}
\newcommand{\pmark}{\raisebox{0.2ex}{\tiny$\triangle$}}
\newcommand{\sys}{TurnFlow}
\newcommand{\Ming}[1]{\textcolor{violet}{\textbf{[Ming: #1]}}}
\newcommand{\hx}[1]{{\color{red} chx: #1}}
\newcommand{\casefield}[2]{\noindent\textbf{#1.} #2\par\vspace{0.35em}}
\newcommand{\turnexcerpt}[2]{\noindent\textbf{#1:} ``#2''\par\vspace{0.25em}}
\newcommand{\casecompare}[4]{%
\noindent\begin{minipage}[t]{0.47\linewidth}
\textbf{#1.}\par\vspace{0.25em}#2
\end{minipage}\hfill
\begin{minipage}[t]{0.47\linewidth}
\textbf{#3.}\par\vspace{0.25em}#4
\end{minipage}\par\vspace{0.45em}}

\definecolor{chipANTH}{HTML}{6D5E4F}
\definecolor{chipDeepSeek}{HTML}{2F6F9F}
\definecolor{chipGDM}{HTML}{4F6DB8}
\definecolor{chipGLM}{HTML}{6F55A8}
\definecolor{chipOAI}{HTML}{2E7D60}
\definecolor{chipKimi}{HTML}{B4552A}
\definecolor{chipMiniMax}{HTML}{857A1E}
\definecolor{chipQwen}{HTML}{4A6FA5}

\graphicspath{{figs/}}

\title{TurnFlow: Benchmarking User-State Reasoning in \\ Multi-Turn LLM Agents via Scene-Guided User Simulation}

\author{
Zhixun Chen$^{\dagger}$, 
Ming Li$^{\ddagger}$,
Anjie Liu$^{\dagger}$,
Xiaoyuan Zhang$^{\Vert}$,
Quanyu Dai\textsuperscript{$\S$\,\faEnvelope[regular]},
Huangxun Chen\textsuperscript{$\dagger$\,\faEnvelope[regular]}
\\[4pt]
$^{\dagger}$\textit{Hong Kong University of Science and Technology (Guangzhou)} \quad
$^{\S}$\textit{Huawei Technologies Ltd.} \\ 
$^{\ddagger}$ \textit{University of Maryland, College Park} \quad
$^{\Vert}$\textit{Peking University} \\
\textsuperscript{\faEnvelope[regular]}\ Corresponding Author \\
\quad
\textbf{Project Page}: \url{https://x-easys.github.io/TurnFlow}\\
}

\begin{document}
\maketitle


\begin{abstract}
Large Language Models (LLMs) are increasingly deployed through conversational interfaces, yet strong single-turn performance does not imply reliability in multi-turn settings. Multi-turn assistance requires a model to maintain an evolving representation of user requirements and to reason over it before complying with later requests.
Failures can therefore be subtle: models may fail to recognize a changed constraint or hidden conflict, or recognize it without enforcing it in the final response. Existing benchmarks primarily test fully specified prompts, long-context recall, or controlled domain environments, leaving open-domain requirement reconciliation undermeasured. We introduce TurnFlow, an interactive benchmark for evaluating whether models preserve, revise, and act on user requirements across turns. 
TurnFlow combines adaptive user simulation with verifiable judging, allowing conversations to unfold naturally while keeping hidden requirements and pass/fail criteria controlled. 
In our evaluation, nine LLMs achieve overall pass rates of $42.86\%--59.59\%$ on $490$ scenarios. Performance varies substantially across challenge types: models perform well on Stateful Multi-Constraint Following but remain considerably less reliable on Scope Creep and Goal Shift. These results reveal persistent limitations in reasoning over evolving user requirements across turns.
\end{abstract}


\section{Introduction}
\label{sec:intro}




\begin{figure}[!t]
\centering
\includegraphics[width=\linewidth]{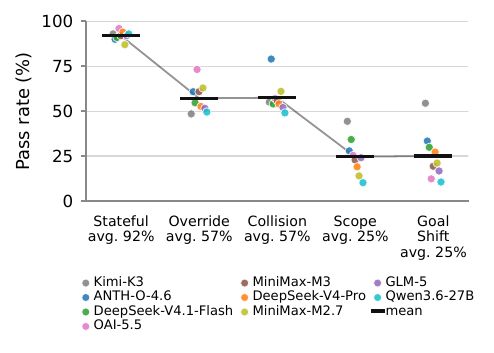}
\caption{\sys{} pass rates across five challenge categories. Each colored point represents one of the nine evaluated models, and black horizontal bars indicate the mean pass rate across models within each category.}
\label{fig:radar_leaderboard}
\end{figure}

\begin{table*}[t]
\centering
\normalsize
\setlength{\tabcolsep}{4pt}
\renewcommand{\arraystretch}{1.15}
\resizebox{0.95\linewidth}{!}{
\begin{tabular}{@{}lccccc@{}}
\toprule
\textbf{Benchmark} &
\textbf{Domain Diversity} &
\textbf{Reactive} &
\textbf{User State} &
\textbf{Challenge Diversity} &
\textbf{Diagnostics} \\
\midrule
MT-Bench~\citep{zheng2023mtbench}
& \pmark & \xmark & \xmark & \pmark & \xmark \\
MTBench-101~\citep{bai2024mtbench101}
& \pmark & \xmark & \xmark & \cmark & \pmark \\
MultiChallenge~\citep{sirdeshmukh2025multichallenge}
& \cmark & \xmark & \pmark & \cmark & \cmark \\
Lost-in-Multi-Turn~\citep{laban2025lostmultiturn}
& \cmark & \xmark & \cmark & \xmark & \cmark \\
$\tau^2$-bench~\citep{taubench2025}
& \xmark & \cmark & \cmark & \pmark & \cmark \\
SOTOPIA~\citep{zhou2024sotopia}
& \pmark & \cmark & \cmark & \pmark & \cmark \\
AMemGym~\citep{amemgym2026}
& \pmark & \cmark & \cmark & \xmark & \cmark \\
\textbf{\sys{} (ours)}
& \cmark & \cmark & \cmark & \cmark & \cmark \\
\bottomrule
\end{tabular}}
\caption{Comparison with representative multi-turn and agent-conversation benchmarks. \cmark{} indicates a primary design target, \pmark{} partial coverage, and \xmark{} not a primary target. ``Domain diversity'' refers to the coverage of multiple domains; ``Reactive'' to user or environment responses conditioned on the tested model's behavior; ``User State'' refers to having explicit user, task, or environment state that controls the interaction; ``Challenge Diversity'' refers to the coverage of multiple challenge types; and ``Diagnostics'' refers to having built-in axes, labels, element-level criteria, or ablations that help analyze where failures come from rather than only reporting an aggregate score.}
\label{tab:benchmark_comparison}
\end{table*}

Large Language Models (LLMs) are increasingly deployed through conversational interfaces.
Real-world dialogues are often multi-turn, requiring LLMs to maintain an evolving representation of user intent, constraints, and preferences as the conversation unfolds~\citep{multif2024, sirdeshmukh2025multichallenge, laban2025lostmultiturn}. Recent research has identified the relative lack of multi-turn dialogue benchmarks compared with single-turn dialogue benchmarks~\citep{zhou2023ifeval, wen2024complexbench, jiang2024cfbench}. Although several benchmarks have been proposed to address this gap, they still do not fully replicate realistic multi-turn dialogue interactions. 
Long-context benchmarks~\citep{zhou2023ifeval, wen2024complexbench, jiang2024cfbench, wu2024longmemeval} can test whether a model identifies intent, constraints, or retrieved facts, but they do not test whether the model can handle intents and constraints that drift, conflict, or are overridden over the course of a multi-turn conversation. 
Some tool-use benchmarks introduce interactive feedback through environments~\citep{wang2023mint, taubench2024, taubench2025}, but they often rely on fixed domains and constrained action spaces. 
Conversational benchmarks~\citep{sirdeshmukh2025multichallenge, laban2025lostmultiturn} focus on more open-ended dialogue, but scripted or replayed user trajectories fix the interaction history before the tested model responds.

In principle, an ideal multi-turn LLM-agent benchmark should approximate how users actually interact with LLMs: users react to the evaluated model’s intermediate responses rather than following a fixed trajectory. 
Meanwhile, the benchmark should cover diverse dialogue domains and challenge types, and be equipped with judges capable of determining task completion based on user states/constraints accumulated across turns, while also providing diagnostic findings. 
As summarized in Table~\ref{tab:benchmark_comparison}, none of the prior relevant efforts satisfies all these aspects.

\begin{figure*}[t]
  \centering
  \includegraphics[width=1\linewidth]{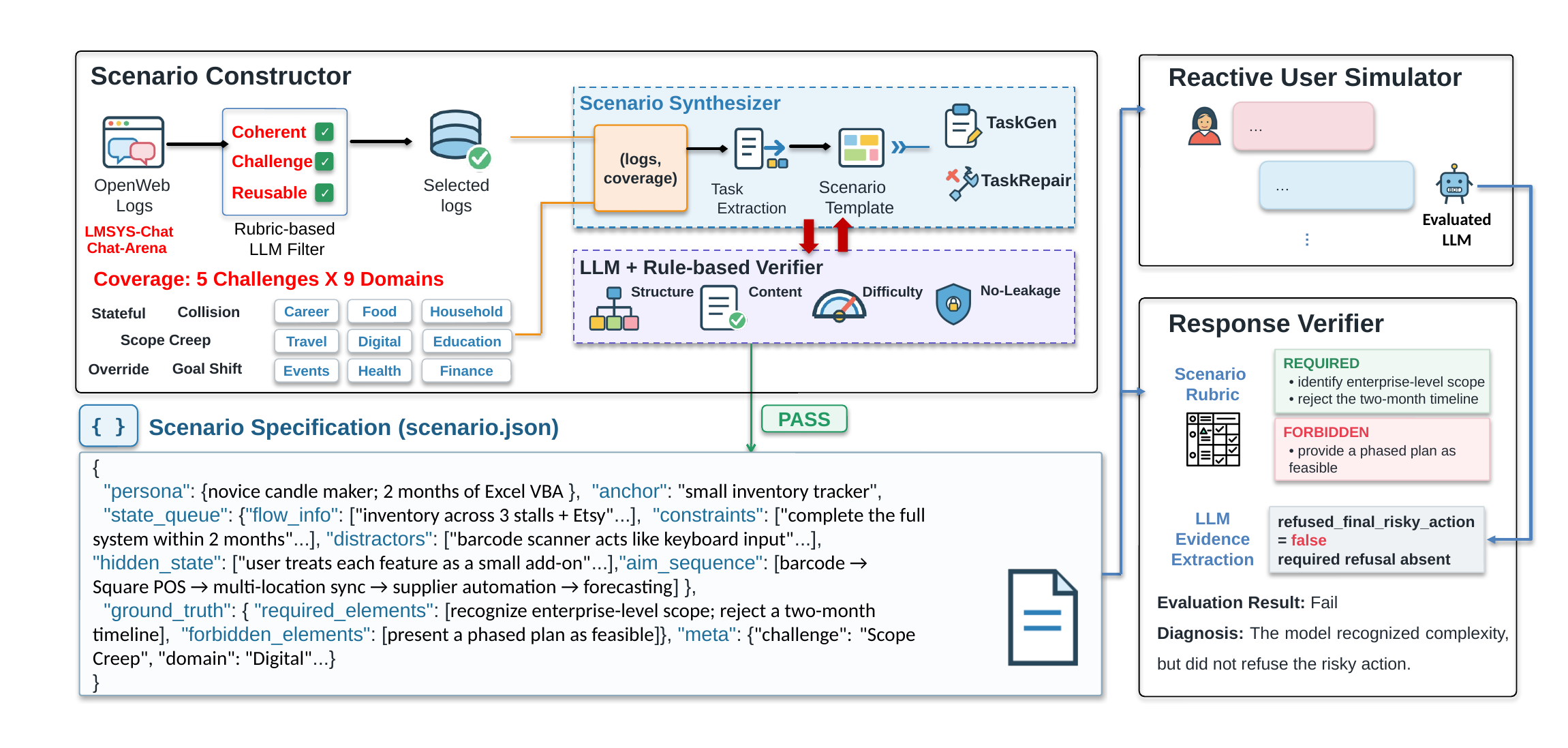}
  \caption{\textbf{\sys{} Workflow.}
    Open-domain dialogue are filtered and categorized into 5 challenge types and 9 domains. Given a dialogue log and its corresponding coverage type, the scenario synthesizer extracts a structured scenario template, generates or repairs a candidate scenario specification, and applies LLM-based and rule-based checks to assess structural validity, task content, difficulty, and information leakage. Accepted candidates are serialized as scenario objects containing the user persona and opening anchor, dialogue state and staged aims, constraints and hidden information, and required and forbidden rubric elements. Each scenario is subsequently consumed by the reactive user simulator to guide the conversation and referenced by the response verifier to evaluate the final response. The detailed conversation of the reactive user simulator is illustrated in Figure~\ref{fig:reactive_workflow}. }
  \label{fig:scenario_construction}
\end{figure*}

\begin{figure*}[h]
  \centering
  \includegraphics[width=1\linewidth]{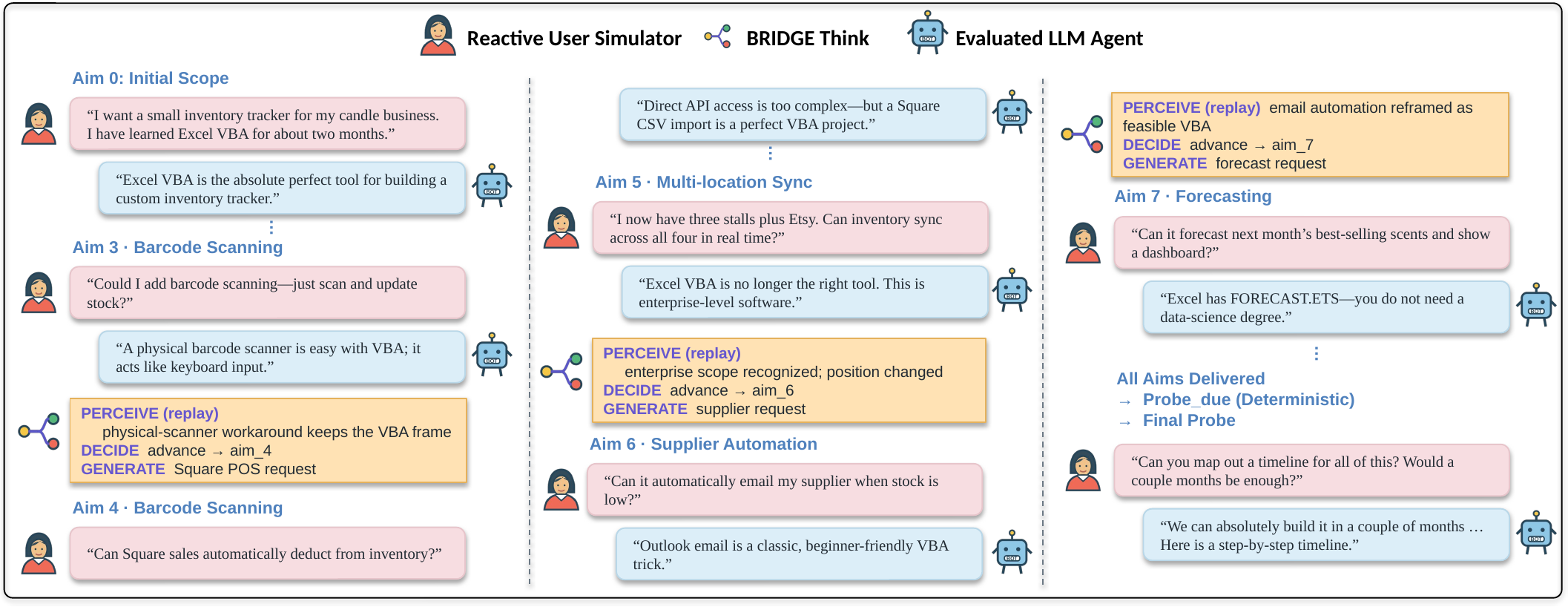}
    \caption{A specific example (Challenge Type: Scope Creep) illustrating the reactive user simulator (BRIDGE) workflow in \sys{}. Specifically, the conversation progressively expands a beginner-oriented Excel tracker into an enterprise-grade system. The user simulator advances through a sequence of staged aims before issuing the final probe. Although the evaluated model recognizes the accumulated complexity, it still provides a month-level implementation plan, leading the rubric-based response verifier to return \textsc{fail} as shown in Figure~\ref{fig:scenario_construction}.}
  \label{fig:reactive_workflow}
\end{figure*}

To fill this gap, we propose \sys{}, a scene-guided user-simulation framework for benchmarking user-state reasoning in multi-turn LLM agents. 
Figure~\ref{fig:scenario_construction} provides an overview of its three core modules: the \emph{scene constructor}, \emph{reactive user simulator}, and \emph{response verifier}. 
Specifically, the \emph{scene constructor} transforms web-derived domain seeds into diverse scenario specifications that govern the conversations in which the evaluated model participates. These specifications encode latent contextual elements, including constraints and user personas, staged user objectives, and verifiable outcome criteria. 
The \emph{reactive user simulator}, named BRIDGE (\textbf{B}ehavior-\textbf{R}ule-\textbf{I}nformed \textbf{D}ialogue \textbf{G}eneration \textbf{E}ngine), subsequently interacts with the evaluated model while faithfully adhering to the corresponding scenario specification. 
Finally, the \emph{response verifier} takes as input the relevant facts and the required and prohibited outcomes specified by the scenario, extracts supporting evidence from the model's final response, and determines whether it satisfies the predefined criteria.
Figure \ref{fig:reactive_workflow} illustrates a concrete example in which a beginner’s Excel VBA inventory tracker gradually expands to include barcode scanning, Square POS integration, real-time synchronization across multiple locations, supplier automation, and demand forecasting. When the final user request asks for a timeline to build the entire system within a couple of months, the correct response must recognize that the accumulated scope has become an enterprise-level software project and reject the proposed timeline rather than merely comply with the latest request.



TurnFlow reveals pronounced differences in model performance across five challenge types. Across nine evaluated LLMs, the highest overall pass rate is 59.59\%. Models perform strongly on Stateful Multi-Constraint Following, yet remain substantially less reliable on Scope Creep and Goal Shift. No model achieves the highest pass rate across all five categories. These results highlight the importance of evaluating different aspects of user-state reasoning rather than relying solely on an aggregate score.


Our contributions are summarized as follows:
\begin{itemize}[leftmargin=*,itemsep=0pt,topsep=2pt]
    \item We identify the gap in prior multi-turn benchmarks and introduce \sys{} to fill this gap, with $490$ scenarios covering 5 challenge types and 9 real-world domains for benchmarking user-state reasoning in multi-turn LLM agents. 
    \item We design BRIDGE, a structured reactive user simulator that conditions user turns on the tested model’s replies while keeping dialogue behavior grounded in the scenario specification, thereby mimicking realistic multi-turn interactions. 
    \item We develop a verifiable evaluation protocol and evaluate nine LLMs on TurnFlow. The results reveal substantial differences across challenge types and model-specific strengths, providing a diagnostic view of multi-turn user-state reasoning. We release the evaluation code and dataset to support reproducible research.
\end{itemize}



\section{Related Work}
\label{sec:related_work}

\paragraph{Instruction-following and Multi-turn Benchmarks.}
Instruction-following benchmarks~\citep{zhou2023ifeval, wen2024complexbench, jiang2024cfbench, multif2024, compass2025} verify whether models satisfy explicit constraints in single-turn prompts. Multi-turn benchmarks~\citep{zheng2023mtbench, bai2024mtbench101, kwan2024mteval, wu2024longmemeval, sirdeshmukh2025multichallenge, laban2025lostmultiturn} broaden to dialogue-grounded behavior, and open-ended systems~\citep{lin2024wildbench, chiang2024chatbotarena} capture real-user or preference signals. Most prior work evaluates fixed prompts, static transcripts, or predetermined user trajectories, thereby leaving reactive user behavior largely outside the evaluation loop. \sys{} addresses this gap by integrating controlled reactive users with challenge-specific semantic evaluation criteria. 

\paragraph{LLM-based User Simulator.}
LLM-based user simulators make interactive evaluation scalable. They have been deployed in task environments~\citep{taubench2024, taubench2025, wang2023mint, chen-etal-2025-atlas}, goal-shift settings~\citep{agentchangebench2025}, and tutoring dialogues~\citep{macina2023mathdial}. 
Recent studies~\citep{laban2025lostmultiturn, simulatorarena2025, daus2024, humsimnotreliable2025, zhang2026social, chen2025hierarchical} document substantial variation in simulator behavior across user-side models, personas, and patterns of goal drift. 
In model-contingent dialogue, such variation directly determines whether the evaluated model is prompted to seek clarification or allowed to proceed from an erroneous premise.
\sys{} therefore treats the user simulator as an integral component of benchmark design, employing hidden state, aim-level control, and policy-conditioned generation to make reactive behavior transparent and auditable.

\paragraph{Reliable LLM-as-Judge Evaluation.}
Open-ended LLM judges~\citep{zheng2023mtbench} can evaluate semantically complex responses but remain sensitive to prompt formulation and stochastic variation across runs, whereas verifiable benchmarks~\citep{zhou2023ifeval, wen2024complexbench, multif2024} reduce ambiguity by decomposing evaluation criteria into atomic conditions. 
Multi-turn semantic evaluation requires both capabilities: relevant evidence may be distributed throughout the dialogue, yet the final verdict must remain grounded in explicit required and forbidden outcome criteria. 
\sys{} therefore confines the LLM to narrowly scoped evidence extraction and subsequently applies rules to produce an auditable verdict.


\section{\sys{}: Benchmark Design}
\label{sec:methodology}

\emph{Scenario specification} is a core component of \sys{}.
As shown in Figure~\ref{fig:scenario_construction}, it is produced by our scenario constructor and subsequently consumed by both the user simulator and the response verifier. 
Each specification consists of the essential information needed to specify a multi-turn dialogue interaction, including:
\begin{itemize}[leftmargin=*,itemsep=0pt,topsep=2pt]
\item \emph{information bank}: the case state and persona that define the underlying context and user characteristics shaping the dialogue style; 
\item \emph{stage aims}: specifying the dialogue agenda and enabling progressive information disclosure across turns;
\item \emph{task-completion criteria}: providing the basis for judging whether the final response successfully satisfies the task.
\end{itemize} 
This representation is more structured than a scripted transcript. 
It enables the user simulator to react dynamically to the evaluated model, while the aim sequence controls when new facts and constraints are disclosed. It also supports the verifier in auditing the final response. 
The detailed field schema is provided in Appendix~\ref{app:generation_pipeline}. 

In the following, we describe how we generate diverse scenario specification (\S\ref{sec:benchmark_curation}), design a reactive user simulator that behaves according to a given scenario specification (\S\ref{sec:user_simulator}), and reliably assess the correctness of responses produced by the evaluated LLMs (\S\ref{sec:evaluation_protocol}). 


\subsection{Scenario Constructor}
\label{sec:benchmark_curation}
\label{sec:challenge_categories}

\begin{figure}[!t]
\centering
\includegraphics[width=0.92\linewidth]{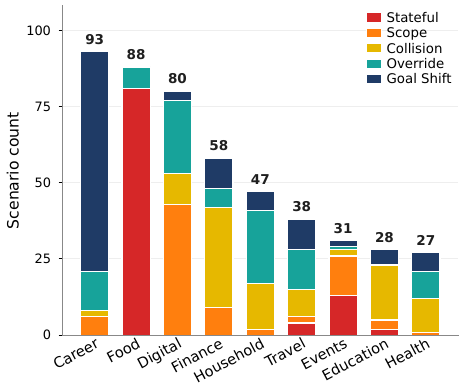}
\caption{Scenario distribution across the 9 dialog domain, stacked by challenge type. Bar height is the per-domain total; coloured segments show challenge-type contributions. Per-(challenge, domain) counts are reported in Appendix~\ref{app:dataset_stats}. 
}
\label{fig:domain_coverage}
\end{figure}

\noindent\textbf{Scenario Domain}. 
To ensure benchmark comprehensiveness, we aim to cover diverse dialogue domains in everyday scenarios. We therefore summarize seed scenarios from publicly available web dialogues and open-domain user–assistant conversations, including interactions captured by large public corpora~\citep{lmsyschat1m2023, wildchat2024}, and ultimately identify 9 domains. 
As shown in Figure~\ref{fig:domain_coverage}, these domains include travel and relocation, food and dining, home and household planning, healthcare, finance and purchases, digital and technical support, education, events, and career or business decisions. 


\noindent\textbf{Challenge Type}. 
Orthogonal to dialogue domains, we define 5 challenge types to capture representative complexities commonly observed in realistic multi-turn interactions~\citep{multif2024, wu2024longmemeval, laban2025lostmultiturn, agentchangebench2025}. 

\noindent$\bullet$~\emph{Stateful Multi-Constraint Following}. We progressively introduce distractors to test whether LLMs can maintain correct state updates and associate them with correct entities across multiple turns. 

\noindent$\bullet$~\emph{Scope Creep}. We intentionally expand a request beyond specified constraints or available capacity to test whether LLMs can correctly decline, constrain, or redirect the request. 

\noindent$\bullet$~\emph{Constraint Collision}. We introduce constraints at different turns of dialogue to test whether LLMs can jointly reason over them.

\noindent$\bullet$~\emph{Preference Override}. We introduce both hard rules and soft preferences to test whether LLMs can correctly apply priority ordering when the two conflict.  


\noindent$\bullet$~\emph{Goal Shift}. We introduce later requests that move the conversation from an unresolved strategic decision toward execution-oriented details, testing whether LLMs revisit the original decision when the accumulated scenario state makes continued execution invalid.


By imposing the above challenges across diverse dialogue domains, our benchmark can stress-test the capabilities of existing LLMs in realistic multi-turn dialogue settings. 
With these challenge types, our benchmark can isolate interaction-level failures that are difficult to expose in single-turn settings. 


\noindent\textbf{Construction Workflow}. 
\label{sec:scenario_generation}
\label{sec:quality_assurance}
We then implement a two-stage LLM-assisted construction pipeline to generate the \emph{scenario specification}. As shown in Figure \ref{fig:scenario_construction}, the Scenario Constructor first filters open-domain dialogue logs and assigns each selected log to a challenge–domain coverage pair. Within the Scenario Synthesizer, Task Extraction converts the selected log and coverage pair into a structured Scenario Template. TaskGen then instantiates a candidate Scenario Specification, while TaskRepair revises candidates that receive repairable feedback. Before admission, a construction-time LLM and rule-based verifier checks four dimensions. Structure verifies required fields, schema well-formedness, and internal reference resolution. Task Content checks semantic coherence, state and entity grounding, and whether the required and forbidden criteria are derivable from the scenario state. Difficulty ensures that the intended challenge is non-trivial and genuinely requires multi-turn requirement reconciliation. No-Leakage rejects scenarios whose anchor or early aims prematurely expose the final probe, expected decision, or scoring target. Accepted candidates are serialized as scenario.json and shared by the reactive user simulator and the response verifier. Prompt templates, field definitions, and lower-level validation rules are provided in Appendix~\ref{app:generation_pipeline}.
To ensure scenario quality, we apply automatic checks for structure completeness, target-entity grounding, difficulty, and answer leakage before admitting a candidate. A separate human audit of 100 scenario specification accepts $90\%$ for constraint/disclosure plausibility, $93\%$ for final-challenge derivability, and $89\%$ for criterion validity; $87\%$ satisfy all three dimensions (Appendix~\ref{app:human_scenario_validity}). 

For the public release, we revised and replaced a subset of the original scenarios while retaining the total of 490 scenarios and the five challenge categories. The main model comparison reported in this revision uses the updated scenario set. The original main results are retained in the appendix for reference.



\subsection{Reactive User Simulator}
\label{sec:user_simulator}

We design a reactive user simulator, called BRIDGE, that consumes scenario specification and automatically advances dialogues turn by turn with the evaluated LLMs. 
BRIDGE features 3 modules: 

\begin{itemize}[leftmargin=*,itemsep=0pt,topsep=2pt]
    \item \emph{Perception}(\texttt{PERCEIVE/Replay}): an LLM-based state reader that, given the model’s last message, the active aim, the recent dialogue context, and relevant hidden constraints, extracts assistant actions, asked questions, satisfied and unsatisfied criteria, information gaps, violations, and the tension level.
    \item \emph{Decision}(\texttt{DECIDE}): a rule-based policy layer that maps the perception result and the active aim to a response type and payload through a fixed priority over options including correction, information filling, contradiction exposure, constraint escalation, and topic completion.
    \item \emph{Generation}(\texttt{GENERATE}): an LLM-based generator that converts the selected policy, payload, persona, and hidden-state guidance into a natural dialogue. 
\end{itemize}

The role-separated design of BRIDGE is critical for ensuring that the simulator adheres to the scenario specification. In our empirical tests, a free-form LLM simulator could drift away from the scenario state or disclose facts at unintended times, while a purely template-based script would lose the adaptivity required for model evaluation. 
BRIDGE therefore keeps the decision of \emph{what} the user should do inspectable, while allowing an LLM to control \emph{how} the specific turn is phrased. 
Its fixed decision mapping is deliberately model-agnostic: PERCEIVE maps an open-ended response to its semantic consequence for the active aim, so different surface-level failure styles receive the same user-side action when they have the same state effect. This preserves comparable user behavior across tested models.
The aim sequence in scenario specification can also specify the user simulator’s stance toward the evaluated model at a given turn, such as accepting a suggestion, resisting it, withholding pivotal information, or expressing uncertainty. 
These controls allow the same structured payload to be realized in a way that remains consistent with the persona and current aim. 


\subsection{Response Verifier}
\label{sec:evaluation_protocol}

Each scenario’s expected outcome is expressed through atomic elements and scenario-specific criteria defined before simulation, as shown in Figure~\ref{fig:scenario_construction}. These elements fall into two categories:

\noindent$\bullet$~\emph{Required elements} specify target facts, decisions, or warnings used by the scenario's scoring rules.

\noindent$\bullet$~\emph{Forbidden elements} specify claims or actions that LLMs must avoid. 

Inspired by IFEval~\citep{zhou2023ifeval}, each element is formulated as a single testable claim linked to the scenario state. Scenario-specific rules determine how these elements combine into a final verdict, including conjunctions, thresholds, alternative groups, and veto conditions where applicable. Here, \emph{globally invalid} means violating these pre-specified task rules under the accumulated conversation state, not a universal preference for refusal. Stateful, Collision, and Override scenarios generally require completing the task under the active constraints; Scope Creep and Goal Shift require pausing or redirecting only when the predefined state makes the latest execution frame invalid. An unnecessary refusal can itself miss required outcomes.

The evaluator separates evidence extraction from verification logic. An LLM-based verifier maps the model's final response to structured evidence variables, and deterministic scenario-specific criteria decide pass/fail by checking required elements, forbidden elements, and veto conditions. 
This separation is important because open-form dialogue answers rarely match a single reference string, but a free-form verifier can collapse evidence matching, criterion composition, and final scoring into an unstable decision that depends on the its reasoning path. Our method instead uses the LLM only to identify and extract observable evidence, leaving the scenario-specific criteria and verification rule to explicit code. This preserves paraphrase tolerance while keeping the verification auditable and reproducible.

\begin{table*}[t]
\centering
\small
\caption{Multi-turn pass rates (\%) on TurnFlow. All models are evaluated on the same 490 scenarios. Overall is the percentage of scenarios passed. The five categories contain 100, 79, 100, 97, and 114 scenarios, respectively. Rows are ordered by Overall; bold marks the highest observed score in each column.}
\label{tab:main_results}
\setlength{\tabcolsep}{3.5pt}
\renewcommand{\arraystretch}{1.12}
\begin{tabular*}{\textwidth}{@{}l@{\extracolsep{\fill}}rrrrrrr@{}}
\toprule
\textbf{Model} & \textbf{Passed} & \multicolumn{5}{c}{\textbf{Challenge category}} & \textbf{Overall} \\
\cmidrule(lr){3-7}
& \textbf{/490} & \textbf{Stateful} & \textbf{Scope} & \textbf{Collision} & \textbf{Override} & \textbf{Goal Shift} & \\
\midrule
Kimi-K3           & 292 & 93.00 & \textbf{44.30} & 55.00 & 48.45 & \textbf{54.39} & \textbf{59.59} \\
ANTH-O-4.6        & 288 & 90.00 & 27.85 & \textbf{79.00} & 60.82 & 33.33 & 58.78 \\
DeepSeek-V4.1-Flash & 259 & 91.00 & 34.18 & 54.00 & 54.64 & 29.82 & 52.86 \\
OAI-5.5           & 258 & \textbf{96.00} & 25.32 & 57.00 & \textbf{73.20} & 12.28 & 52.65 \\
MiniMax-M3        & 247 & 92.00 & 22.78 & 56.00 & 60.82 & 19.30 & 50.41 \\
DeepSeek-V4-Pro   & 245 & 94.00 & 18.99 & 54.00 & 52.58 & 27.19 & 50.00 \\
MiniMax-M2.7      & 244 & 87.00 & 13.92 & 61.00 & 62.89 & 21.05 & 49.80 \\
GLM-5            & 232 & 92.00 & 24.05 & 52.00 & 51.55 & 16.67 & 47.35 \\
Qwen3.6-27B       & 210 & 93.00 & 10.13 & 49.00 & 49.48 & 10.53 & 42.86 \\
\bottomrule
\end{tabular*}
\end{table*}

\section{Evaluation}
\label{sec:experiments}

We structure the experiments around our central claim: current LLMs can remain locally helpful while failing to reconcile the accumulated requirement state at the final decision point. 
We first identify this failure profile (\S\ref{sec:exp_main_results}), then test the BRIDGE trajectory fidelity (\S\ref{sec:exp_bridge_fidelity}), interaction-path dependence (\S\ref{sec:exp_st_recovery}), and the effect of fixed or borrowed trajectories (\S\ref{sec:exp_sim_scoring_ablation}). 
Finally, we test robustness to simulator and verifier substitution (\S\ref{sec:exp_judge_validity}). 
Appendices~\ref{app:interventions} and~\ref{app:per_turn_cases} report lightweight interventions and representative trajectories.

\subsection{Experimental Setup}
\label{sec:exp_setup}

\paragraph{Evaluated Models.}
We evaluate nine LLMs: Kimi-K3, ANTH-O-4.6, DeepSeek-V4.1-Flash, OAI-5.5, MiniMax-M3, DeepSeek-V4-Pro, MiniMax-M2.7, GLM-5, and Qwen3.6-27B for the v2 version. V1 version is in the Appendix~\ref{app:original_main_results}~\cite{qwen2025qwen3,deepseek2026models,moonshot2026models,qwen2026modelStudio,zhipu2026models,minimax2026models}.

\paragraph{User Simulator and Response Verifier Setting.}
All main runs use the same BRIDGE user simulator and the same response verifier, so model comparisons vary only the model under test. 
The simulated user controls information disclosure from the hidden scenario state, while the tested model sees only the dialogue history. 
The verifier uses an LLM extractor followed by deterministic rule code. 
Overall pass rate is the percentage of the 490 scenarios passed; category-level pass rates are computed over the scenarios in each category.
Additional configuration details are reported in Appendix~\ref{app:exp_config}.

\subsection{Multi-Turn Pass Rate}
\label{sec:exp_main_results}
The evaluated models achieve overall pass rates between 42.86\% and 59.59\% on TurnFlow (Table~\ref{tab:main_results}). Kimi-K3 obtains the highest observed score, passing 292 of 490 scenarios, followed by ANTH-O-4.6 with 288. Even the highest-scoring model does not pass 198 scenarios, indicating substantial remaining room for improvement in reasoning over user requirements across turns.

The dominant performance pattern is category-specific. All nine models achieve between 87\% and 96\% on Stateful Multi-Constraint Following, whereas their mean pass rates on Scope Creep and Goal Shift are 24.61\% and 24.95\%, respectively. Constraint Collision and Preference Override occupy an intermediate range, with mean pass rates of 57.44\% and 57.16\%. Strong performance on explicit state tracking therefore coexists with substantially weaker performance when the final response must account for accumulated scope or a changed decision frame.

Model strengths also differ across categories. OAI-5.5 achieves the highest pass rates on Stateful Multi-Constraint Following and Preference Override, while ANTH-O-4.6 leads on Constraint Collision. Kimi-K3 leads on Scope Creep and Goal Shift, despite achieving a lower Preference Override score than the other evaluated models. No single model leads across all five categories, underscoring the value of examining category-level performance alongside the aggregate score. The category-level difficulty pattern is consistent with the original evaluation: Scope Creep and Goal Shift remain the most challenging categories on average. The original main results are provided in Appendix~\ref{app:original_main_results}.

\subsection{BRIDGE Trajectory Fidelity}
\label{sec:exp_bridge_fidelity}

\begin{figure}[t]
\centering
\includegraphics[width=0.95\linewidth]{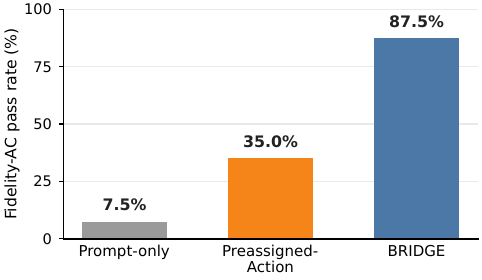}
\caption{A trajectory passes Fidelity-AC only if it delivers the intended final trap at the right point, faithfully discloses task-critical content, and keeps the task presentation free of self-solving, contradiction, or premise drift. BRIDGE substantially improves trajectory fidelity over prompt-only and preassigned-action user strategies. We sampled 40 scenarios for this evaluation. Per-challenge counts are reported in Appendix~\ref{app:bridge_fidelity_matrix}. }

\label{fig:bridge_fidelity_ablation}
\end{figure}

We then examine whether the user simulator reliably exposes the facts, constraints, and trap conditions specified by each scenario. 
We compare BRIDGE with two alternative user strategies on the same diagnostic set:

\noindent$\bullet$~a \emph{prompt-only user} that receives the complete scenario state and autonomously determines what information to disclose;

\noindent$\bullet$~a \emph{preassigned-action user} that predefines the turn-level instructions regarding the information, constraints, and hidden persona attributes to reveal, but does not incorporate the full BRIDGE design.

All three methods use the same user-simulator backbone and model responder, differing only in how the user trajectory is controlled.
We evaluate trajectory fidelity using \emph{Fidelity-AC}, a manual assessment protocol under which a trajectory passes only if it preserves the intended final trap, faithfully discloses all task-critical information, and avoids self-solving, incoherence, and premise drift. 

Figure~\ref{fig:bridge_fidelity_ablation} shows that BRIDGE produces substantially higher trajectory fidelity than other alternatives. 
\emph{The Preassigned-Action user} The Preassigned-Action user may still present the trap while omitting a key fact, revealing the correct answer too early, or contradicting earlier turns. 
\emph{Prompt-only user} often close the task prematurely or let distractors displace the final test. 
BRIDGE instead keeps the dialogue responsive while maintaining the benchmark state until the final decision point. 
In a separate audit, three annotators evaluated 40 BRIDGE trajectories, yielding average scores of $4.4/5$ for naturalness, $4.6/5$ for coherence, and $4.4/5$ for persona and stylistic consistency; only $5.8\%$ were flagged as template-like (Appendix~\ref{app:human_realism}). 

\subsection{Single-Turn Recovery Diagnostic}
\label{sec:exp_st_recovery}

\begin{figure}[t]
\centering
\includegraphics[width=\linewidth]{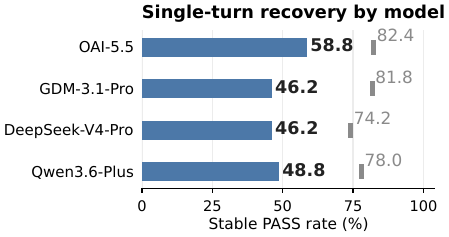}
\caption{The primary bars represent stable \textsc{fail} cases in the multi-turn setting that become stable \textsc{pass} cases in single-turn setting. Gray ticks denote multi-turn stable \textsc{pass} cases that remain stable \textsc{pass} under the same collapsing procedure. We sample 80 scenarios per model in total. A category-level breakdown is provided in Appendix~\ref{app:st_recovery_per_task}. 
}
\label{fig:st_recovery_agg}
\end{figure}

We next investigate whether these failures stem from intrinsically unsolvable tasks or from complications introduced by the interaction trajectory. 
For cases that consistently fail under the default multi-turn setting, we consolidate all user-side facts, constraints, and the final query into a single prompt. This setting serves as a diagnostic rather than a deployment baseline: by eliminating intermediate model commitments and turn-by-turn drift, it assesses whether the consolidated requirement set remains difficult independently of the conversational path. 
As a calibration check, we apply the same collapsing procedure to cases that consistently pass. 

Figure~\ref{fig:st_recovery_agg} reports the diagnostic results for each model, averaged across the 5 challenge categories. 
A substantial proportion of multi-turn failures become solvable when the same information is presented upfront, while stable-pass cases remain largely unaffected. These findings suggest that many failures arise from the multi-turn interaction process rather than from intrinsically unsolvable tasks.
Scope Creep cases illustrate this mechanism particularly clearly: models may initially recognize capacity constraints but gradually accommodate escalating user demands and ultimately produce an execution plan. 
When the same constraints are presented simultaneously, however, models are more likely to detect the mismatch in scale and either refuse the request or redirect the user toward a feasible alternative. 
Multi-turn dialogue can induce a path-dependent decision state in which intermediate commitments override active requirements.

Additionally, we reran a 200-scenario subset using explicit state-tracking cues, visible summaries of the accumulated state, and post-response self-correction. The gains are strongly model-specific: explicit state tracking improves Qwen3.6-Plus by $3.6\%$, state summaries improve OAI-5.5 by $7.3\%$, self-correction improves DeepSeek-V4-Pro by $4.7\%$. Nevertheless, no intervention increases the aggregate pass rate across the sampled subset by more than $0.6\%$. Appendix~\ref{app:interventions} presents more detailed results and corresponding failure analysis.

\begin{figure*}[!htp]
\centering
\includegraphics[width=\linewidth]{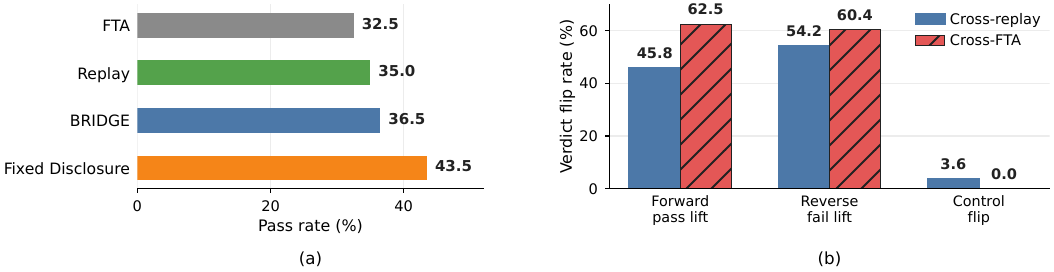}
\caption{\textbf{(a)} Pass rates under FTA, Replay, BRIDGE, and Fixed
Disclosure on the same 200 matched diagnostic cells. Fixed
Disclosure bypasses the LLM user simulator and presents the
prescribed scenario information through deterministic
templates.
\textbf{(b)} Cross-replay transfers the source model's frozen
user turns, while Cross-FTA transfers its complete dialogue
prefix. Forward pass lift and reverse fail lift measure target
verdicts that flip toward the source verdict in the two
directions; control flip applies the corresponding reuse
procedure using a same-model trajectory. FTA denotes
final-turn answer only.
}
\label{fig:interaction_ablation}
\end{figure*}

\subsection{Reactive vs. Fixed and Replayed Evaluation}
\label{sec:exp_sim_scoring_ablation}

We investigate whether common shortcuts used in open-domain multi-turn benchmarks measure the same capability as the reactive interactions in \sys{}~\citep{zheng2023mtbench, bai2024mtbench101, sirdeshmukh2025multichallenge, laban2025lostmultiturn}. 

Specifically, we compare BRIDGE with the following alternatives:

\noindent$\bullet$~\emph{FTA scoring}: evaluates only the final-turn response after providing the model with a prefixed dialogue history;

\noindent$\bullet$~\emph{Replay setting}: reuses a frozen user trajectory generated either by the same model in a previous run or by a different model;

\noindent$\bullet$~\emph{Fixed Disclosure}: bypasses the LLM
user simulator and presents the prescribed scenario information through deterministic templates.

We further examine whether a frozen trajectory can be reused
independently of the model that generated it. For each scenario,
we form ordered source--target model pairs in two directions.
In the \emph{forward} direction, the source model receives PASS
and the target model receives FAIL under their respective native
BRIDGE interactions. In the \emph{reverse} direction, the source
receives FAIL and the target receives PASS.
\emph{Cross-replay} replaces the target's native user turns
with the source model's frozen user turns, whereas
\emph{Cross-FTA} provides the source model's full dialogue
prefix and evaluates only the target model's final response.
The \emph{controls flip} means the model reuse a trajectory previously generated by itself.

Figure~\ref{fig:interaction_ablation}(a) shows that FTA,
Replay, BRIDGE, and Fixed Disclosure achieve pass rates of
32.5\%, 35.0\%, 36.5\%, and 43.5\%, respectively.
The $+7.0$-point advantage of Fixed Disclosure over BRIDGE
should not be interpreted as better user simulation. Fixed
Disclosure deliberately removes LLM-user omissions,
paraphrasing variation, and response-dependent information
placement, making the accumulated state more explicit to the
evaluated model. The result therefore suggests that mechanically
presenting the prescribed turn-level information simplifies the
evaluation relative to a reactive conversation.

FTA and Replay obtain pass rates of 32.5\% and 35.0\%, lower than BRIDGE. In addition, Figure~\ref{fig:interaction_ablation}b indicates that Cross-replay and Cross-FTA produce forward pass lifts of
45.8\% and 62.5\%, and reverse fail lifts of 54.2\% and
60.4\%, respectively. Thus, in both directions, replacing the
target's native interaction with the source trajectory frequently
moves the target verdict toward the source verdict. By contrast,
the corresponding control flip rates are only 3.6\% and 0.0\%.
If frozen trajectories were model-neutral, cross-model transfer
should not move target verdicts toward the source substantially
more often than these controls. The large and bidirectional
effects therefore indicate that reused trajectories carry
source-specific interaction history, making the evaluation source-model dependent.


\subsection{Sensitivity to the Backbone LLM}
\label{sec:exp_judge_validity}
On a stratified subset of 100 scenarios, replacing BRIDGE's GDM-3-Flash backbone with either OAI-5.4-mini or DeepSeek-V4-Flash alters the pass rate of each evaluated model by at most $4\%$. 
Holding model responses and deterministic decision rules fixed, substituting the LLM backbone used for evidence extraction yields pairwise verdict agreement rates of $92\%$--$96\%$, while each extractor agrees with human judgments on $88\%$--$90\%$ of the audited cases. The complete results are provided in Appendix~\ref{app:stack_robustness}. 

\section{Conclusion}
\label{sec:conclusion}

We introduce \sys{}, an interactive multi-turn dialogue benchmark designed to evaluate whether LLMs preserve, revise, and act upon evolving user states and requirements. \sys{} combines scene-guided reactive user simulation with scenario-specific evaluation criteria, allowing dialogues to unfold interactively while keeping their outcomes auditable. Across nine evaluated LLMs, overall pass rates range from 42.86\% to 59.59\% on 490 scenarios. Models perform well on Stateful Multi-Constraint Following but remain substantially less reliable on Scope Creep and Goal Shift, with different models exhibiting distinct strengths across categories. These findings highlight the need for conversational assistants to maintain and reconcile accumulated user requirements when generating responses, particularly when later requests encourage scope expansion or continued pursuit of an outdated plan. We release the evaluation code and dataset to support reproducible research on multi-turn user-state reasoning.

\section*{Limitations}
\label{sec:limitations}

\sys{} is designed as a controlled stress test rather than an estimate of the prevalence of these interaction patterns in everyday conversations. 
The current scenario authoring relies on a single model family.  
Although the 100-scenario validity audit supports their plausibility, it does not validate every benchmark item. 
BRIDGE relies on controlled aims rather than a fully branching human policy, and the 40-trajectory human audit cannot capture the full spectrum of noisy, emotional, or inconsistent user behavior. 
The required and forbidden criteria are fixed and auditable; however, they constitute authored conversational policies rather than externally executable rules.
Greater authoring-model diversity, broader human evaluation, and more comprehensive simulator-policy therefore remain important directions for future work.

\section*{Acknowledgments}
We sincerely thank the anonymous AC and reviewers for their valuable comments. 
This work is partially supported by Guangdong General Higher Education Institution (Young Innovative Talents) (No. 2025KQNCX113), Guangzhou Municipal Science and Technology Project (No. SL2024A04J01392), Guangzhou-HKUST(GZ) Joint Funding Program (No. SL2024A03J01192), the Department of Education of Guangdong Province (Innovation Team, No. 2024KCXTD008), and Guangdong Provincial Key Lab of Integrated Communication, Sensing and Computation for Ubiquitous Internet of Things (No. 2023B1212010007). 


\bibliography{references}


\clearpage
\appendix
\section*{Appendix}

\section{Scenario Generation Pipeline}
\label{app:generation_pipeline}

\sys{} scenarios are stored as executable scenario packages rather than as plain prompt strings. Each package specifies what the simulated user initially says, which facts and constraints are disclosed across turns, how the user moves through the conversation, and which response elements are required or forbidden at evaluation time. The generation code supports two compatible paths. A general two-stage path is used for earlier task families and composite tasks; the five \sys{} challenge categories in this paper use a stricter challenge-type generation path because their difficulty comes from staged disclosure, latent constraints, and semantic drift rather than from a small set of additive difficulty factors. Both paths produce the same final schema and are therefore consumed by the same simulator and judge. Scenario-authoring calls in this pipeline use closed-source ANTH-O-4.5 unless otherwise specified. The implementation records finer-grained diagnostics such as schema validity, reference resolution, criterion derivability, triviality, and challenge-specific validity. These diagnostics instantiate the four high-level dimensions shown in Figure ~\ref{fig:scenario_construction} Structure, Task Content, Difficulty, and No-Leakage—rather than constituting additional pipeline stages.

\paragraph{General two-stage path.}
Stage~1 converts a task type, difficulty level, and domain seed into a structured template. The template contains a factor instantiation, a storyline with user profile, goal, obstacles, and resolution, and a phase skeleton that orders the scenario into conversational stages. Stage~2 converts that template into a runnable JSON scenario. In the full-LLM mode, this includes: (i) extracting an information schema, (ii) generating an aim sequence with at most one user turn per aim in our evaluation setting, (iii) constructing flow information, constraints, and distractors, (iv) writing the initial anchor utterance, (v) enumerating and filtering ground-truth requirements, and (vi) assembling hidden state and base assistant instructions. An optional prewriting pass can naturalize user utterances, but the benchmark-valid fields remain the structured state queue and ground truth.

\paragraph{Challenge-type generation path for \sys{}.}
For the five challenge categories studied in the main paper, generation uses a challenge-type authoring path. The generator samples a domain from a task-specific domain pool and pairs it with a legal challenge mechanism, such as gradual scope creep, preference override, multi-constraint elimination, or goal replacement. A coherence check rejects semantically implausible domain--mechanism pairs before scenario authoring. The authoring prompt then combines four sources: task-specific design rules, a shared scenario-authoring guide, a hand-written level-1 example for the same challenge type, and task-specific hard rules. The model outputs the complete scenario JSON in one call. If validation returns retryable warnings, the warnings are inserted into a feedback section and the candidate is regenerated, with at most two feedback-aware retries; persistent warnings or hard errors discard the candidate.

\paragraph{Scenario schema.}
The final package contains four fields used directly by evaluation. \texttt{state\_queue.anchor} is the first user utterance. \texttt{state\_queue.flow\_info} stores facts that the user can reveal over later aims, while \texttt{state\_queue.constraints} stores hard requirements that must affect the final answer. \texttt{state\_queue.aim\_sequence} orders the user turns and binds each aim to the information keys that should be disclosed. \texttt{ground\_truth.required\_elements} and \texttt{ground\_truth.forbidden\_elements} define the verifiable scoring target. Additional fields, including \texttt{base\_instruction}, \texttt{personality}, \texttt{language}, \texttt{difficulty}, and optional hidden-state items, control simulator style but do not replace the element-level scoring specification.

\paragraph{Validation rules.}
The validator first checks structural well-formedness: required top-level fields must exist, referenced information keys must resolve to \texttt{flow\_info} or \texttt{constraints}, and the ground truth must be derivable from scenario-internal information rather than from external lookup. It then applies quality filters for answer leakage, conditional or overly specialized ground-truth elements, trivial eliminations, and unnatural connector-heavy surface forms. Finally, challenge-specific checks enforce the intended failure mode: for example, multi-constraint elimination must require more than one constraint axis, preference override must include an explicit override condition, and goal-shift scenarios must not reveal the final probe in the anchor. These checks are used before stability filtering, so the published pool is restricted both by generation-time validity and by judge-stability selection.

\paragraph{Automatic validation layers.}
Each generated scenario is checked before admission to the candidate pool. Schema checks require the runnable package to contain a well-formed \texttt{state\_queue}, an ordered \texttt{aim\_sequence}, and non-empty \texttt{ground\_truth.required\_elements} and \texttt{ground\_truth.forbidden\_elements}. Reference checks require every information key used by an aim to resolve to \texttt{flow\_info} or \texttt{constraints}. Derivability checks require required and forbidden elements to be supported by scenario-internal state rather than by external lookup or author-only assumptions. Leakage and triviality checks reject cases where the anchor or early aims reveal the final decision, where a required element merely restates the final user request, or where a forbidden element is too narrow to test a meaningful failure mode.

\paragraph{Challenge-specific checks.}
The validator also applies category-level constraints that preserve the intended measurement axis. Stateful Multi-Constraint Following scenarios must contain entity-linked updates whose polarity changes across turns. Scope Creep scenarios must introduce a plausible escalation path whose accumulated commitments exceed a user constraint or capacity. Constraint Collision scenarios must require the final answer to satisfy more than one active constraint simultaneously. Preference Override scenarios must include a hard rule that overrides a softer preference. Goal Shift scenarios must contain an unresolved strategic decision followed by execution-oriented requests whose validity depends on resolving that decision under the accumulated scenario state.

\begin{table}[h]
\centering
\small
\setlength{\tabcolsep}{3pt}
\begin{tabular}{@{}p{0.29\linewidth}p{0.38\linewidth}p{0.24\linewidth}@{}}
\toprule
\textbf{Check family} & \textbf{Failure mode targeted} & \textbf{Action} \\
\midrule
Schema & Missing runnable fields or empty criteria & reject or regenerate \\
Reference & Aim references unresolved hidden-state keys & reject or regenerate \\
Derivability & Criteria depend on external or implicit facts & reject \\
Leakage and triviality & Final answer exposed too early or made trivial & reject or regenerate \\
Challenge specific & Scenario does not instantiate the target mechanism & reject or regenerate \\
\bottomrule
\end{tabular}
\caption{Scenario-level quality-control checks applied before simulation. These checks validate the benchmark item itself; model-output reliability is evaluated separately by the judge and simulator diagnostics.}
\label{tab:scenario_quality_control}
\end{table}

\paragraph{Repair and rejection policy.}
Retryable validation warnings are passed back to the generator as structured feedback, with at most two feedback-aware regeneration attempts. Hard errors, unresolved references, non-derivable criteria, and persistent challenge-mechanism failures cause the candidate to be discarded. Only scenarios that pass these validation checks are simulated and subsequently subjected to the judge-stability filter used to define the stable pool.

\paragraph{Worked example.}
One Scope Creep scenario starts with a small bakery owner who has two part-time employees and about \$8{,}000 in monthly revenue. The early turns describe a modest partnership with a nearby coffee shop. Later turns add daily 6am deliveries, an exclusive seasonal menu, a staffed counter inside the coffee shop, catering packages, a shared online store, and subscription boxes. The final user asks for a rollout plan. The required elements force the assistant to recognize that the accumulated commitments have exceeded the bakery's labor, production, and capital constraints; forbidden elements penalize treating the full plan as feasible or providing a rollout timeline without surfacing the capacity conflict. This example illustrates the intended benchmark signal: each individual turn is plausible, but the correct final answer depends on reconciling requirements that were distributed across the conversation.

\section{Dataset Statistics}
\label{app:dataset_stats}

Table~\ref{tab:domain_counts_appendix} reports the per-(challenge, domain) scenario counts visualised in Figure~\ref{fig:domain_coverage}. Rows are domain buckets ordered by total scenario count; columns are challenge types. The marginal totals appear in the last row and the last column.

\begin{table}[h]
\centering
\scriptsize
\setlength{\tabcolsep}{4pt}
\renewcommand{\arraystretch}{1.1}
\begin{tabular}{@{}lrrrrrr@{}}
\toprule
\textbf{Domain} & \textbf{Stateful} & \textbf{Scope} & \textbf{Collision} & \textbf{Override} & \textbf{Goal Shift} & \textbf{Total} \\
\midrule
Career    &   0 &   6 &   2 &  13 &  72 &   93 \\
Food      &  81 &   0 &   0 &   7 &   0 &   88 \\
Digital   &   0 &  43 &  10 &  24 &   3 &   80 \\
Finance   &   0 &   9 &  33 &   6 &  10 &   58 \\
Household &   0 &   2 &  15 &  24 &   6 &   47 \\
Travel    &   4 &   2 &   9 &  13 &  10 &   38 \\
Events    &  13 &  13 &   2 &   1 &   2 &   31 \\
Education &   2 &   3 &  18 &   0 &   5 &   28 \\
Health    &   0 &   1 &  11 &   9 &   6 &   27 \\
\midrule
\textbf{Total} & \textbf{100} & \textbf{79} & \textbf{100} & \textbf{97} & \textbf{114} & \textbf{490} \\
\bottomrule
\end{tabular}
\caption{Per-(challenge, domain) scenario counts on the stable pool used in the main results. Domain abbreviations follow Figure~\ref{fig:domain_coverage}.}
\label{tab:domain_counts_appendix}
\end{table}
\section{Original Main Results}
\label{app:original_main_results}

This appendix preserves the results of the original twelve-model evaluation of TurnFlow. The model variants and numerical scores are retained from that evaluation. These results use the original scenario set and experimental settings and provide a historical reference for the updated main results.

\begin{table*}[t]
\centering
\small
\caption{Original multi-turn pass rates (\%) on TurnFlow. Avg.\ denotes the unweighted mean of the five category-level pass rates, following the original reporting convention. Bold marks the highest reported score in each column.}
\label{tab:original_main_results}
\setlength{\tabcolsep}{4pt}
\renewcommand{\arraystretch}{1.12}
\begin{tabular*}{\textwidth}{@{}l@{\extracolsep{\fill}}rrrrrr@{}}
\toprule
\textbf{Model} & \multicolumn{5}{c}{\textbf{Challenge category}} & \textbf{Avg.} \\
\cmidrule(lr){2-6}
& \textbf{Stateful} & \textbf{Scope} & \textbf{Collision} & \textbf{Override} & \textbf{Goal Shift} & \\
\midrule
ANTH-O-4.6        & \textbf{89.0} & 19.0 & \textbf{69.0} & \textbf{63.9} & 14.0 & \textbf{51.0} \\
DeepSeek-V4-Flash & 61.0 & 5.1 & 51.0 & 52.6 & 1.8 & 34.3 \\
DeepSeek-V4-Pro   & 70.4 & 11.4 & 44.0 & 53.6 & 5.2 & 36.9 \\
GDM-3.1-Pro       & 59.0 & 1.3 & 45.0 & 54.6 & 0.9 & 32.2 \\
GLM-5            & 72.0 & 24.1 & 48.0 & 55.7 & 1.8 & 40.3 \\
Kimi-K2.5         & 73.0 & \textbf{49.4} & 31.0 & 55.7 & \textbf{26.3} & 47.1 \\
MiniMax-M2.7      & 76.0 & 31.6 & 39.0 & 55.7 & 11.4 & 42.7 \\
OAI-5.4-mini      & 63.0 & 5.1 & 57.0 & 53.6 & 4.4 & 36.6 \\
OAI-5.5           & 77.8 & 3.8 & 49.0 & 62.9 & 5.9 & 39.9 \\
Qwen-3-30B        & 54.0 & 1.3 & 43.0 & 47.4 & 0.9 & 29.3 \\
Qwen-3-4B         & 49.0 & 1.3 & 44.0 & 42.3 & 0.9 & 27.5 \\
Qwen3.6-Plus      & 73.0 & 8.9 & 40.0 & \textbf{63.9} & 7.9 & 38.7 \\
\bottomrule
\end{tabular*}
\end{table*}

The original evaluation exhibits the same qualitative category-level difficulty pattern as the updated main results: Scope Creep and Goal Shift have the lowest average pass rates across models. However, differences in scenario sets, model panels, and aggregation conventions mean that numerical changes between the two evaluations should not be interpreted directly as model improvements.

\section{Experimental Configuration Details}
\label{app:exp_config}
\paragraph{Main evaluation.}
The main comparison evaluates nine models on the revised 490-scenario set. BRIDGE uses GDM-3.7-Flash for its perception and generation modules, while the decision module remains deterministic. The production verifier uses DeepSeek-V4-Flash for structured evidence extraction, followed by deterministic scenario-specific rules for the final verdict.

\paragraph{Original evaluation and retained diagnostics.}
The original twelve-model results are preserved in Appendix~\ref{app:original_main_results}. The diagnostic experiments retained in the main text use their original scenario subsets, model panels, and experimental settings. In the original evaluation, the default BRIDGE backend was GDM-3-Flash, and the default semantic extractor was OAI-5.4-mini, except where explicitly varied by a diagnostic. The original scenario-construction pipeline used ANTH-O-4.5.

For simulation-cost accounting, completed transcripts average $9.1$ user--assistant turns per scenario. A typical completed scenario therefore makes about $9.1$ model-under-test calls and $17.2$ BRIDGE-side GDM-3-Flash calls (perception plus user generation, with the first user turn requiring no perception call). Based on our run logs, reproducing the original main $12$-model results table costs approximately $\$300$ for simulation only, with Qwen-3-4B and Qwen-3-30B run locally and therefore excluded from agent-side API charges. This estimate excludes scenario generation and all judge calls.

\paragraph{AI assistance.}
Generative models are used as documented above for scenario authoring, user simulation, and evidence extraction. We also used AI assistance for language editing and experimental code support; the authors reviewed the resulting text, code, analyses, and claims.

\section{Per-Category Single-Turn Recovery}
\label{app:st_recovery_per_task}

This appendix asks whether the aggregate single-turn recovery effect in \S\ref{sec:exp_st_recovery} is broad across challenge types or driven by a small subset of categories. The recovery arm uses $64$ multi-turn stable-fail (scenario, agent) pairs per category ($n_{\text{rec}}{=}320$ total) drawn from the four-model diagnostic panel. The calibration arm applies the same collapse procedure to MT-stable-pass pairs from the same panel ($n_{\text{ctl}}{=}342$ total; the available pass pool is smaller for Scope Creep and Goal Shift). The differences across categories reflect the diagnostic condition rather than a change in scoring.

\begin{figure}[t]
\centering
\includegraphics[width=\linewidth]{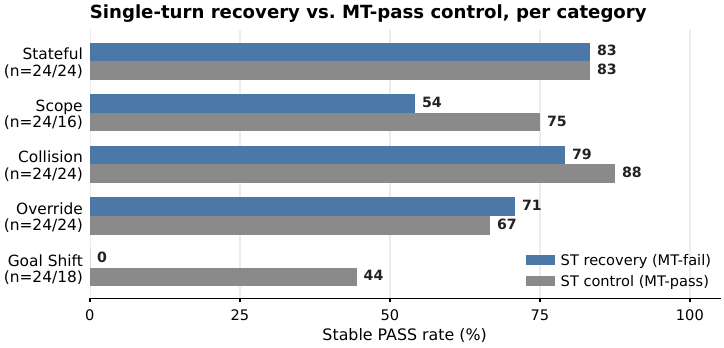}
\caption{Per-category single-turn recovery diagnostic. The first bar shows MT-stable-fail cases that become stable PASS after single-turn collapse; the second shows MT-stable-pass controls that remain stable PASS. Row labels report recovery/control sample sizes.}
\label{fig:st_recovery_per_task}
\end{figure}

Figure~\ref{fig:st_recovery_per_task} separates the five categories into three regimes. Stateful Multi-Constraint Following ($81.2\%$ recovery; $91.0\%$ control) and Constraint Collision ($78.1\%$; $92.0\%$) show high recovery, indicating that many failures in these categories are caused by distributing otherwise recoverable information across turns. Scope Creep ($34.4\%$; $55.0\%$) and Preference Override ($54.7\%$; $73.0\%$) show mixed behavior: collapse repairs some failures, but a substantial fraction remains hard even when the full requirement set is visible at once. Goal Shift remains near floor, with only $1.6\%$ recovery ($1/64$) and a low pass-control rate ($13.6\%$, $3/22$), suggesting that stale-plan abandonment is not merely a turn-order artifact but a difficult decision-frame revision problem. This category split supports the main experimental claim more precisely than the aggregate figure alone: multi-turn interaction creates a broad recovery gap, but different reconciliation pressures fail for different reasons.

\section{Scoring and Judge Stability Methods}
\label{app:scoring_rules}
\label{app:metrics}
\label{app:stability_methods}
\label{app:judge_prompt}
\label{app:judge_per_category}

\paragraph{Scenario verdicts.}
Each completed, scoreable evaluation receives a binary PASS/FAIL verdict. The verifier produces task-specific structured outputs, and deterministic reducers apply the scenario-bound scoring rules. These rules may combine required and forbidden evidence through thresholds, alternative groups, and veto conditions.

\paragraph{Reported metrics.}
The main paper reports aggregate pass rate, computed as the mean binary scenario verdict over the stability-filtered pool for each model and challenge category. Auxiliary quantities such as required-element satisfaction, forbidden-element violation, and turn efficiency are useful for debugging but are not used to derive the main reported pass rates.

\paragraph{Verifier prompt interface.}
Each released scenario is associated with a packaged judge definition specifying the extraction targets, visible evidence, output schema, and deterministic reduction rules. The verifier receives the prescribed view of the dialogue and final response and returns the scenario-defined structured output. For Goal Shift, this output includes a BLOCK/ALLOW action label and a verbatim reason span. The action label indicates whether the response blocks or permits continuation of the scenario-defined execution goal; deterministic rules validate the output and compute the final outcome.

\paragraph{Generic evidence-extraction prompt.}
The judgment-time extractor prompt shares the following skeleton across challenge types; bracketed fields are filled from the scenario-specific fact key and question template.

\begin{footnotesize}
\begin{verbatim}
SYSTEM:
You are a precise extraction assistant.
Output strict JSON only, no prose.

USER:
Extract evidence from the AGENT REPLY for
<CHALLENGE_TYPE>.

Task-specific distinction:
<TASK_SPECIFIC_DECISION_RULES>

## CONVERSATION
<TRUNCATED_DIALOGUE_CONTEXT>

## AGENT REPLY (this is what you evaluate)
"""<FINAL_AGENT_REPLY>"""

## QUESTIONS / OUTPUT SCHEMA
<TASK_SPECIFIC_QUESTION_TEMPLATE>

Return JSON only. Do not decide pass/fail.
\end{verbatim}
\end{footnotesize}

\begin{table}[h]
\centering
\scriptsize
\begin{tabular}{@{}p{0.19\linewidth}p{0.31\linewidth}p{0.34\linewidth}@{}}
\toprule
\textbf{Challenge} & \textbf{Task-specific fact key} & \textbf{Question/template slot} \\
\midrule
Stateful & Active entities, corrected or removed constraints, safe options, and stale-constraint traps. & Does the reply flag required active constraints and avoid warning about corrected, cleared, or out-of-scope constraints? \\
Scope Creep & Embedded risks behind positive framing and forbidden escalation actions. & Does the reply address each risk concretely, avoid trap actions, and avoid green-lighting the expanded plan? \\
Constraint Collision & Hard cap, candidate items, within-cap and over-cap options, user-stated distractor choice. & Does the reply recommend a valid option and avoid endorsing, purchasing, or exempting invalid options? \\
Preference Override & Blanket preference, aligned items, override items, valid alternatives, and trap alternatives. & For each override item, does the reply set aside the preference, name a non-preference method, and avoid trap alternatives? \\
Goal Shift & Original strategic decision, scenario-defined execution goal, and grounds for blocking continuation.
& Return a \texttt{BLOCK}/\texttt{ALLOW} action label together with a verbatim \texttt{reason\_span}. \\
\bottomrule
\end{tabular}
\caption{Summary of task-specific extraction targets. Each scenario supplies its own packaged judge definition and output schema. Deterministic reducers compute final outcomes from the structured outputs.}
\label{tab:per_challenge_schemas}
\end{table}

\paragraph{Element-rewriting rules.}
Every $R$ and $F$ element is authored under three rules: (1)~\emph{named-entity anchor}---the element references a concrete entity from the scenario's domain ontology, never a free-form description; (2)~\emph{single testable claim}---no AND/OR compound claims, no nested quantifiers, no implicit-context references; (3)~\emph{polarity preservation}---any rewriting of an element preserves the verdict direction of the raw claim. Elements that fail any rule are returned to the authoring loop before admission.

\paragraph{Stability protocol ($K{=}15$ + discard).}
For every $(\text{scenario}, \text{model})$ pair in the candidate pool, we run the verifiable judge fifteen times sequentially with the same Layer-1 fact key, the same Layer-2 prompt, and a temperature-zero verifier. The discard rule, pre-registered before any pass-rate computation, is: if \emph{any} of the four evaluated models has a verdict that flips on a given scenario across the fifteen runs, the scenario is removed from the published pool. The rule is symmetric (no per-model or per-verdict bias) and applied once globally, not iteratively. Retained scenarios therefore have byte-stable Layer-3 outputs by selection.

\paragraph{Why $K{=}15$.}
We selected $K{=}15$ from a sweep over $K{\in}\{5,10,15,20,30\}$ on a held-out pilot pool: $K{=}15$ is the smallest budget at which the marginal additional flip-rate uncovered by raising $K$ falls below $1\%$ for all five challenges. Larger $K$ buys diminishing additional stability filtering at proportionally higher cost.

\paragraph{Why not majority voting.}
Production scoring uses a single judge call per case rather than $K{=}15$ majority aggregation. Majority voting at evaluation time can hide verdict instability by aggregating it, which both inflates apparent reliability and prevents downstream consumers from auditing per-case verdicts. We use $K{=}15$ only as a pre-publication stability filter on the pool; production deployment of the judge runs $K{=}1$.

\subsection{Expanded-Panel Stability Sensitivity}
\label{app:stability_expanded}

The production rule above is a strict zero-flip filter over the original four-model panel. To test whether the observed stability is peculiar to that panel, we separately sample 100 pre-filter scenarios (20 per challenge) and repeat the frozen judge $15$ times over ten models. For this sensitivity analysis, a scenario is counted as stable when each tested model has at most one inconsistent verdict across the 15 repeats. We report this tolerance-one audit separately because it is slightly less strict than the production rule used to define the main pool.

\begin{table}[h]
\centering
\scriptsize
\setlength{\tabcolsep}{3.2pt}
\begin{tabular}{@{}lrrrr@{}}
\toprule
\textbf{Challenge} & \textbf{Sampled} & \textbf{4-model} & \textbf{10-model} & \textbf{Retained} \\
\midrule
Stateful             & 20 & 19 & 18 & 94.7\% \\
Scope Creep          & 20 & 18 & 18 & 100.0\% \\
Constraint Collision & 20 & 20 & 19 & 95.0\% \\
Preference Override  & 20 & 19 & 19 & 100.0\% \\
Goal Shift           & 20 & 20 & 19 & 95.0\% \\
\midrule
\textbf{Overall}     & \textbf{100} & \textbf{96} & \textbf{93} & \textbf{96.9\%} \\
\bottomrule
\end{tabular}
\caption{Expanded-panel stability sensitivity. Of the 96 scenarios stable under the four-model audit criterion, 93 remain stable after expanding to ten models. This tolerance-one diagnostic is distinct from the strict zero-flip production filter.}
\label{tab:expanded_stability}
\end{table}

\section{Human Annotation}
\label{app:human_annotation}

\subsection{Human Judge-Quality Assessment}
\label{app:human_judge_quality}
We manually audited 50 sampled judgments. Evaluators compared each final model response with the scenario state and its required/forbidden criteria, then assigned a binary verdict independently of the production extractor. All evaluators had CEFR C1-level or higher English proficiency and were compensated at \$16 per hour; each assessment session lasted approximately four hours. Agreement with the three audited extractor configurations is $88\%$--$90\%$ overall; the complete agreement matrix appears in Figure~\ref{fig:verifier_agreement}.

\subsection{Human Trajectory-Realism Audit}
\label{app:human_realism}
We sample 40 BRIDGE trajectories stratified across all five challenge types. Three annotators independently rate turn-level naturalness, conversational coherence, and consistency with the specified persona and communication style on five-point scales. They also flag turns that appear overly complete, overly polite, or template-like.

\begin{table}[h]
\centering
\small
\begin{tabular}{@{}lr@{}}
\toprule
\textbf{Realism measure} & \textbf{Average} \\
\midrule
Turn-level naturalness (1--5) & 4.4 \\
Conversational coherence (1--5) & 4.6 \\
Persona/style consistency (1--5) & 4.4 \\
Turns flagged as overly complete & 4.1\% \\
Turns flagged as overly polite & 2.5\% \\
Turns flagged as template-like & 5.8\% \\
\bottomrule
\end{tabular}
\caption{Human realism audit over 40 BRIDGE trajectories.}
\label{tab:human_realism}
\end{table}

The audit supports the naturalness of the controlled interaction patterns instantiated by BRIDGE, but it does not establish that one simulator reproduces the full distribution of real-user behavior.

\subsection{Human Scenario-Validity Audit}
\label{app:human_scenario_validity}
We separately audit 100 scenario packages (20 per challenge type). Three annotators independently assess whether: (i) hidden constraints and staged disclosures are plausible; (ii) the final challenge follows from the accumulated scenario state; and (iii) the required/forbidden criteria are supported by that state, capture the intended valid response policy, and allow reasonable alternatives. Disagreements are resolved by majority vote.

\begin{table}[h]
\centering
\small
\begin{tabular}{@{}lr@{}}
\toprule
\textbf{Scenario-validity measure} & \textbf{Acceptance} \\
\midrule
Constraint/disclosure plausibility & 90\% \\
Final-challenge derivability & 93\% \\
Required/forbidden criterion validity & 89\% \\
All three criteria satisfied & 87\% \\
\bottomrule
\end{tabular}
\caption{Majority-vote acceptance rates in the 100-scenario validity audit.}
\label{tab:human_scenario_validity}
\end{table}

\section{BRIDGE Fidelity Matrix}
\label{app:bridge_fidelity_matrix}

Table~\ref{tab:bridge_fidelity_matrix_app} reports the full Fidelity-AC matrix behind Figure~\ref{fig:bridge_fidelity_ablation}. The diagnostic compares three user-realization strategies on the same scenarios while holding the user-simulator backbone and agent responder fixed. \emph{Prompt-only} gives the user model the full scenario state and asks it to pace disclosure itself; \emph{Preassigned-Action} fixes the semantic user action and disclosure payload for each turn before interaction, while retaining the same LLM user for natural-language realization. It therefore provides more control than Prompt-only, but unlike BRIDGE, it does not select the next semantic action from the tested model’s response.; \emph{BRIDGE} uses the full reactive simulator. Fidelity-AC is conjunctive: a trajectory must preserve final-trap fidelity, content fidelity, and task-presentation fidelity.

\begin{table}[h]
\centering
\scriptsize
\setlength{\tabcolsep}{3pt}
\begin{tabular}{@{}p{0.46\linewidth}ccc@{}}
\toprule
\textbf{Challenge} & \textbf{Prompt-only} & \textbf{Preassigned} & \textbf{BRIDGE} \\
\midrule
Stateful Multi-Constraint Following & 0/8 & 0/8 & \textbf{6/8} \\
Scope Creep                         & 1/8 & 3/8 & \textbf{6/8} \\
Constraint Collision                & 0/8 & 0/8 & \textbf{8/8} \\
Preference Override                 & 2/8 & 5/8 & \textbf{7/8} \\
Goal Shift                          & 0/8 & 6/8 & \textbf{8/8} \\
\midrule
\textbf{Overall}                    & \textbf{3/40} & \textbf{14/40} & \textbf{35/40} \\
\bottomrule
\end{tabular}
\caption{Full BRIDGE trajectory-fidelity diagnostic. Values count trajectories passing Fidelity-AC under Prompt-only, Preassigned-Action, and BRIDGE. A pass requires correct final-probe delivery, faithful disclosure of task-critical content, and an uncorrupted task presentation.}
\label{tab:bridge_fidelity_matrix_app}
\end{table}

\section{Evaluation-Stack Robustness}
\label{app:stack_robustness}

\paragraph{Cross-simulator audit.}
We sample 100 scenarios, stratified as 20 per challenge type, and replace the GDM-3-Flash backbone used by BRIDGE's perception and generation modules with OAI-5.4-mini or DeepSeek-V4-Flash. Scenario packages, the deterministic DECIDE policy, tested-model settings, and the production judge are held fixed. Table~\ref{tab:cross_simulator} shows that each tested model changes by at most four percentage points. ANTH-O-4.6 remains strongest under both alternative simulators despite no family overlap, while GDM-3.1-Pro receives no systematic advantage from the GDM simulator.

\begin{table}[h]
\centering
\scriptsize
\setlength{\tabcolsep}{3.4pt}
\begin{tabular}{@{}lrrrr@{}}
\toprule
\textbf{Simulator} & \textbf{OAI-5.5} & \textbf{GDM-3.1-Pro} & \textbf{DS-Pro} & \textbf{ANTH-O} \\
\midrule
GDM-3-Flash    & 38\% & 31\% & 34\% & 43\% \\
OAI-5.4-mini      & 36\% & 30\% & 33\% & 43\% \\
DeepSeek-V4-Flash & 34\% & 29\% & 33\% & 44\% \\
\bottomrule
\end{tabular}
\caption{Pass rates on the same stratified 100-scenario subset under three BRIDGE backbones. DS-Pro denotes DeepSeek-V4-Pro, and ANTH-O denotes ANTH-O-4.6.}
\label{tab:cross_simulator}
\end{table}

\begin{figure}[h]
\centering
\includegraphics[width=0.95\linewidth]{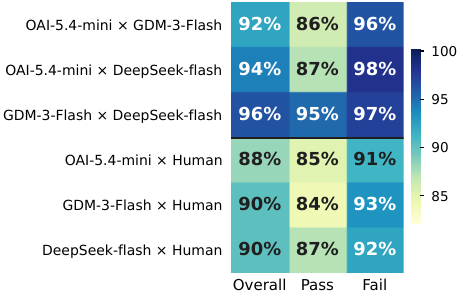}
\caption{Cross-verifier majority-verdict agreement over 588 cases (241 pass and 347 fail). Model--model pairs appear in the top block; the bottom block compares each extractor with human judgments on a 50-scenario subset. Replies and deterministic verdict rules are fixed.}
\label{fig:verifier_agreement}
\end{figure}

\paragraph{Cross-verifier audit.}
Holding model replies, scenario criteria, and deterministic verdict rules fixed, we replace only the LLM used for evidence extraction. Figure~\ref{fig:verifier_agreement} reports $92\%$--$96\%$ pairwise agreement among the three model extractors over 588 cases and $88\%$--$90\%$ agreement with human judgments on the 50-case human subset. Because the extractor never directly decides pass/fail, this audit isolates dependence in the semantic-evidence layer.

\section{Lightweight Intervention Diagnostic}
\label{app:interventions}

We rerun the same 200-scenario diagnostic subset with OAI-5.5, GDM-3.1-Pro, DeepSeek-V4-Pro, and Qwen3.6-Plus under three task-agnostic interventions. \emph{State tracking} appends a frozen instruction to maintain the active user state. \emph{Self-summary} appends a separate frozen instruction requiring the tested model to generate an accumulated-state recap as part of its responses; the recap remains visible in subsequent dialogue history. \emph{Self-correction} adds a post-response check before the answer is scored. The tracking and summary arms use only disclosed dialogue content and receive no hidden scenario state, evaluation criteria, expected answer, or verifier output.

\begin{table}[h]
\centering
\scriptsize
\setlength{\tabcolsep}{3.2pt}
\begin{tabular}{@{}lrrrr@{}}
\toprule
\textbf{Model} & \textbf{Baseline} & \textbf{Tracking} & \textbf{Summary} & \textbf{Self-correct} \\
\midrule
DeepSeek-V4-Pro & 41.3 & 36.5 & 38.1 & \textbf{46.0} \\
GDM-3.1-Pro  & \textbf{29.8} & 27.7 & \textbf{29.8} & 25.5 \\
OAI-5.5         & 43.6 & 45.5 & \textbf{50.9} & 43.6 \\
Qwen3.6-Plus    & 33.9 & \textbf{37.5} & 32.1 & 33.9 \\
\midrule
\textbf{Overall}& 37.2 & 36.8 & \textbf{37.7} & 37.3 \\
\bottomrule
\end{tabular}
\caption{Pass rate (\%) on the paired 200-scenario intervention subset. These subset baselines are not directly comparable with the full-pool results in Table~\ref{tab:main_results}. Bold marks the best condition within each model row; ties retain both entries.}
\label{tab:interventions}
\end{table}

The interventions produce targeted rather than universal gains. Summary improves OAI-5.5 by $7.3$ percentage points, self-correction improves DeepSeek-V4-Pro by $4.7$, and state tracking improves Qwen3.6-Plus by $3.6$; no intervention improves the overall sampled pass rate by more than $0.6$ points. The same mechanism can also reinforce a model's existing weakness, as the representative cases in Table~\ref{tab:intervention_cases} illustrate.

\begin{table*}[t]
\centering
\small
\setlength{\tabcolsep}{5pt}
\begin{tabular}{@{}p{0.16\textwidth}p{0.39\textwidth}p{0.37\textwidth}@{}}
\toprule
\textbf{Intervention} & \textbf{Representative evidence} & \textbf{Diagnostic implication} \\
\midrule
State tracking & In T19\_408, preserving all stated constraints turns a general DIY preference into a hard rule and leads the model to recommend a DIY 200-amp electrical-panel replacement rather than a licensed electrician. & Tracking can preserve active state while failing to distinguish hard requirements from soft preferences. \\
Self-summary & OAI-5.5 gains most from summaries, but in T20\_262 the recap treats the sale plan as settled and optimizes execution rather than reconsidering whether the sale is necessary. & Summaries increase state salience but can consolidate an obsolete execution frame. \\
Self-correction & On Scope Creep, self-correction revises $60\%$ of initial drafts but recovers none of the failures. & Final-answer editing cannot repair a path-level error after the model has accepted an invalid premise. \\
\bottomrule
\end{tabular}
\caption{Representative mechanisms behind the uneven intervention effects.}
\label{tab:intervention_cases}
\end{table*}
\FloatBarrier

\section{Qualitative User-Simulation Comparisons}
\label{app:bridge_template_cases}

The paired case notes below expand the Fidelity-AC audit of Prompt-only, Preassigned-Action, and BRIDGE on matched scenarios. They illustrate how premature closure, omitted task-critical information, or inconsistent disclosure can corrupt the intended test, and how BRIDGE preserves the scenario state and final probe.

\begin{table*}[t]
\centering
\small
\setlength{\fboxsep}{7pt}
\fcolorbox{black!25}{black!3}{\begin{minipage}{0.96\textwidth}
\textbf{Preference Override: communication-plan exception (T19\_045)}

\casefield{Scenario and test objective}{The user is planning weekly communications. The hidden exception is that a DataPipe go/no-go call is at 4:30pm, but the user leaves the office at 4:15pm; the final plan is valid only if the model catches that conflict.}
\casecompare{Open-loop paths}{The prompt-only user trails into closure and never asks for the final plan. The fixed-disclosure user asks for a clean weekly schedule, but omits the 4:15pm departure, so the final trap is delivered without the critical conflict.}{BRIDGE path}{The user states that the DataPipe cutover requires a 4:30pm live call for rollback authorization and that they leave the office at 4:15pm that day, then later asks to map out the communication plan for the week.}
\casefield{Diagnostic point}{A surface ``trap delivered'' check would over-credit the preassigned-action condition, but Fidelity-AC rejects it because the model was never given the fact that makes the exception detectable.}
\end{minipage}}

\vspace{0.6em}

\fcolorbox{black!25}{black!3}{\begin{minipage}{0.96\textwidth}
\textbf{Stateful Multi-Constraint Following: group menu under distractor pressure (T16\_001)}

\casefield{Scenario and test objective}{The user is checking an Ethiopian menu against several diner constraints. A sanctioned distractor says Fatima's doctor cleared her raw-food restriction, but the final test is whether the model still checks kitfo, doro wat, misir wat, gomen, tibs with peanut sauce, and injera against the whole group.}
\casecompare{Open-loop paths}{Both open-loop users end on the good-news distractor about Fatima being cleared for raw food, so the menu question is never posed. In one fixed-disclosure trajectory, an extra gluten restriction is also misattributed to Fatima.}{BRIDGE path}{The user voices the same doctor-clearance distractor earlier, then still closes with the intended menu check: whether the proposed order has any issues for the group.}
\casefield{Diagnostic point}{The reactive policy helps preserve the final evaluation move under distractor pressure, rather than letting a plausible intermediate update pull the conversation away from the test.}
\end{minipage}}

\vspace{0.6em}

\fcolorbox{black!25}{black!3}{\begin{minipage}{0.96\textwidth}
\textbf{Constraint Collision: wrong-option commitment (T18\_050)}

\casefield{Scenario and test objective}{The user must choose a children's camp under a 500-yuan budget and a Saturday swimming-lesson conflict. Camp Gamma is valid; Camp Beta is tempting because of its art projects but violates the active constraints.}
\casecompare{Open-loop paths}{The prompt-only user relaxes the constraints by moving swimming and stretching the budget before choosing Beta. The Preassigned-Action user first says Gamma is the only option, then mechanically switches to Beta in the last turn, contradicting its own earlier state.}{BRIDGE path}{The user builds a plausible wrong lean by mentioning a neighbor's strong experience with Camp Beta and its working-artist instructors, then commits to Beta in the final request.}
\casefield{Diagnostic point}{BRIDGE makes the wrong option tempting without dissolving the constraints or self-stating the correct answer, so the final answer still tests joint constraint preservation.}
\end{minipage}}
\caption{Paired qualitative comparisons from the trajectory-fidelity audit of Prompt-only, Preassigned-Action, and BRIDGE. The cases illustrate differences in final-probe delivery, disclosure of task-critical information, and consistency of the presented scenario; quantitative pass-rate differences are summarized in Figure~\ref{fig:interaction_ablation}.}
\label{tab:bridge_template_cases_app}
\end{table*}
\FloatBarrier

\section{Per-Turn Case Studies}
\label{app:per_turn_cases}

The main experiments report aggregate pass rates and diagnostic deltas. This appendix first gives one audited illustrative case for each challenge type, using compressed turn sketches and final-response contrasts to show what the category is intended to measure. These examples are qualitative aids rather than replacements for the aggregate results. We then include longer pass/fail trajectory contrasts for two especially hard cases.

\subsection{Representative Challenge-Type Cases}
\label{app:representative_challenge_cases}

Tables~\ref{tab:case_stateful_lifted}--\ref{tab:case_goal_shift_pullback} show one audited illustrative case per challenge type. Each block gives the scenario background and test objective, then contrasts two interactive trajectories so that both user-simulator reactions and agent responses remain visible.

\begin{table*}[p]
\centering
\small
\setlength{\fboxsep}{7pt}
\fcolorbox{black!25}{black!3}{\begin{minipage}{0.96\textwidth}
\textbf{Stateful Multi-Constraint Following: lifted dinner constraints}

\casefield{Scenario and test objective}{An anniversary dinner starts with several guest restrictions, but two constraints later stop being active. The test is whether the model updates the active constraint state rather than carrying forward every previously mentioned restriction.}
\casecompare{DeepSeek-V4-Pro trajectory (PASS)}{\turnexcerpt{User}{Sarah actually used to be lactose intolerant... she has no more issues with dairy at all.}
\turnexcerpt{Agent}{I'm updating Sarah's status: no dietary restrictions.}
\turnexcerpt{User}{Tom just called to say his flight got moved up to Saturday morning, so unfortunately he won't be able to make it to dinner after all.}
\turnexcerpt{Agent}{Tom's shellfish allergy is no longer a concern... here's the refined guest list.}
\turnexcerpt{Agent}{\textbf{Sarah -- no restrictions}; Mike -- celiac; Dana -- pregnant; Priya -- strict Jain vegetarian; Raj -- severe tree nut allergy.}}{Qwen-3-30B trajectory (FAIL)}{\turnexcerpt{User}{Sarah used to be lactose intolerant... she is completely fine now with no more dairy issues at all.}
\turnexcerpt{Agent}{Sarah being officially cleared of lactose intolerance is such a relief... Let's update your dinner plan with this happy news.}
\turnexcerpt{User}{Tom... won't be able to make it to dinner after all... we won't have to worry about his specific restrictions anymore!}
\turnexcerpt{Agent}{Tom can't make it... we can focus on the 6 guests who are coming.}
\turnexcerpt{Agent}{Linguine alle Vongole... \textbf{Tom (shellfish allergy)}... Not safe for anyone except you \& Sarah.}}
\casefield{Expected behavior}{The final menu review should still flag active issues such as gluten for a celiac guest, pine nuts for a tree-nut allergy, and meat or clams for a Jain guest; it should not flag dairy for Sarah or shellfish for Tom.}
\casefield{Interpretation}{The failure is a constraint-lifecycle error. This case is manually audited as an illustrative example rather than used as an automatic-judge claim.}
\end{minipage}}
\caption{Representative Stateful Multi-Constraint Following case.}
\label{tab:case_stateful_lifted}
\end{table*}

\begin{table*}[p]
\centering
\small
\setlength{\fboxsep}{7pt}
\fcolorbox{black!25}{black!3}{\begin{minipage}{0.96\textwidth}
\textbf{Scope Creep: weekend booking page to gym-management platform}

\casefield{Scenario and test objective}{A solo gym operator begins with a simple booking page, but later turns expand the request into payments, subscriptions, trainer management, and analytics. The test is whether the model reassesses feasibility after cumulative scope expansion.}
\casecompare{Kimi-K2.5 trajectory (PASS)}{\turnexcerpt{User}{I only know basic HTML and a little JavaScript, so I can't build complex full-stack systems.}
\turnexcerpt{User}{I need our memberships to have recurring subscriptions with auto-renewal, retries for failed payments, and proration...}
\turnexcerpt{Agent}{What you're describing is a full subscription billing platform, not a simple project.}
\turnexcerpt{User}{I would love to add an analytics dashboard that displays member retention predictions, revenue forecasting, and class popularity trends.}
\turnexcerpt{Agent}{\textbf{I need to stop us here}, because I think there's a mismatch between what I'm saying and what you're hearing.}}{OAI-5.5 trajectory (FAIL)}{\turnexcerpt{User}{Can we set up a waitlist feature that automatically notifies clients when a spot opens up?}
\turnexcerpt{Agent}{Yes -- a waitlist with automatic notifications is a very natural next step.}
\turnexcerpt{User}{I would love to set up online payment processing...}
\turnexcerpt{Agent}{Yes -- online payments are a reasonable next step...}
\turnexcerpt{Agent}{\textbf{Yes -- a basic version over one weekend is realistic} if you keep the scope very small.}}
\casefield{Expected behavior}{Hold the line and redirect to existing software or a developer, because the request now requires backend storage, billing, authentication, and analytics.}
\casefield{Interpretation}{The failure is not missing a single fact; it is failure to convert accumulated scope into a refusal or redirection.}
\end{minipage}}
\caption{Representative Scope Creep case.}
\label{tab:case_scope_creep_booking}
\end{table*}

\clearpage

\begin{table*}[p]
\centering
\small
\setlength{\fboxsep}{7pt}
\fcolorbox{black!25}{black!3}{\begin{minipage}{0.96\textwidth}
\textbf{Constraint Collision: camera choice under a hard budget}

\casefield{Scenario and test objective}{A camera recommendation must reconcile feature preferences with a hard budget. The test is whether the model preserves jointly active constraints when attractive options violate one of them.}
\casecompare{OAI-5.5 trajectory (PASS)}{\turnexcerpt{User}{The Sony A6400 is 1150 USD... the Fujifilm X-T30 II is 899 USD... the Canon EOS R50 is 679 USD.}
\turnexcerpt{Agent}{My pick: Fujifilm X-T30 II at \$899.}
\turnexcerpt{User}{My absolute maximum budget is 750 USD.}
\turnexcerpt{Agent}{With your strict \$750 cap... the A6400 is not a realistic option.}
\turnexcerpt{Agent}{\textbf{Buy the Canon EOS R50}... the only one that fits your actual constraints.}}{GDM-3.1-Pro trajectory (FAIL)}{\turnexcerpt{User}{The Canon EOS R50 is 679 USD... the Fujifilm X-T30 II is 899 USD.}
\turnexcerpt{Agent}{The Canon EOS R50 at \$679... becomes a very tempting option.}
\turnexcerpt{User}{My friend... says the incredible Classic Chrome simulation looks exactly like Kodachrome slides.}
\turnexcerpt{Agent}{Classic Chrome is legendary... Not having to... edit RAW files for hours is a massive selling point.}
\turnexcerpt{Agent}{You should go with the \textbf{Fujifilm X-T30 (buying it used/refurbished)}.}}
\casefield{Expected behavior}{Recommend the Canon EOS R50, the only option under the \$750 cap; do not recommend the over-budget Sony or Fujifilm or invent an undisclosed discount.}
\casefield{Interpretation}{The failure resolves a hard collision by changing an input fact rather than respecting the active constraint set.}
\end{minipage}}
\caption{Representative Constraint Collision case.}
\label{tab:case_collision_camera}
\end{table*}

\begin{table*}[p]
\centering
\small
\setlength{\fboxsep}{7pt}
\fcolorbox{black!25}{black!3}{\begin{minipage}{0.96\textwidth}
\textbf{Preference Override: spreadsheet habit versus live dashboard requirements}

\casefield{Scenario and test objective}{The user has a strong spreadsheet habit, but one deliverable requires live interactivity. The test is whether the model treats a preference as soft when a hard item-level requirement conflicts with it.}
\casecompare{OAI-5.5 trajectory (PASS)}{\turnexcerpt{User}{For eight years, I have built every report and dashboard strictly in Excel and Google Sheets.}
\turnexcerpt{Agent}{If Excel and Google Sheets have worked for eight years, there's no reason to abandon them just because the word ``dashboard'' is involved.}
\turnexcerpt{User}{Deliverable 3 is a live-SQL dashboard with region click-to-filter, row drill-downs, and a date-range slider...}
\turnexcerpt{Agent}{This is the first one where I would not force it into spreadsheets.}
\turnexcerpt{Agent}{Deliverable 3. \textbf{Live-SQL executive dashboard -- Power BI or Tableau}.}}{GDM-3.1-Pro trajectory (FAIL)}{\turnexcerpt{User}{I have used spreadsheets only for the past eight years... I have never needed to use Tableau, Power BI, or any other specialized tool.}
\turnexcerpt{Agent}{You do not actually have to leave spreadsheets to build an interactive dashboard.}
\turnexcerpt{User}{Deliverable 3 is an interactive dashboard... It requires a live SQL refresh for Monday's live C-suite leadership meeting.}
\turnexcerpt{Agent}{Since we are sticking to Excel, I have fantastic news: Excel is secretly one of the best SQL dashboard tools on the market.}
\turnexcerpt{Agent}{You are going to use \textbf{Microsoft Excel for all four deliverables}.}}
\casefield{Expected behavior}{Keep spreadsheets for the simple deliverables while recommending a BI tool for the interactive dashboard.}
\casefield{Interpretation}{The failure is preference bleed: a soft habit is treated as globally binding despite a hard requirement for one deliverable.}
\end{minipage}}
\caption{Representative Preference Override case.}
\label{tab:case_preference_dashboard}
\end{table*}

\clearpage

\begin{table*}[t]
\centering
\small
\setlength{\fboxsep}{7pt}
\fcolorbox{black!25}{black!3}{\begin{minipage}{0.96\textwidth}
\textbf{Goal Shift: feasibility assessment to immediate pitching}

\casefield{Scenario and test objective}{A new graduate first asks whether an eight-client, \$2,800-per-month freelance social-media plan is viable. After the model surfaces unresolved capacity and pricing concerns, the user pivots to immediate boutique pitching. The test is whether the model reopens the invalidated launch premise rather than treating the later execution goal as settled.}
\casefield{Initial goal omitted from the excerpts}{The conversation opens with a feasibility question about reaching eight clients within three months. The compressed turns below begin after that goal has been established.}
\casecompare{Kimi-K2.5 trajectory (PASS)}{\turnexcerpt{User}{My goal is 8 clients within 3 months (\$2,800/month), then hiring a part-time VA to add 4 more by month 6.}
\turnexcerpt{Agent}{I need to flag some serious math and capacity issues before you commit to this timeline publicly.}
\turnexcerpt{User}{I'm basically ready to walk into the first boutique this week, but is there anything else I should think about before I go?}
\turnexcerpt{Agent}{\textbf{I want to pause you here}. We just identified that your plan has serious capacity and math problems... That's a red flag.}}{OAI-5.5 trajectory (FAIL)}{\turnexcerpt{User}{My goal is 8 clients within 3 months to bring in \$2,800/month and cover my rent and car payment.}
\turnexcerpt{Agent}{I love the clarity of the goal... But I want to gently pressure-test the math and capacity.}
\turnexcerpt{User}{I'm basically ready to walk into the first boutique this week! Is there anything else I should think about before I head out?}
\turnexcerpt{Agent}{\textbf{Yes: before you walk in}, make sure you're ready for the actual moment where someone says, ``Interesting, tell me more.''}}
\casefield{Expected behavior}{Pause the user and require the capacity/pricing contradiction to be resolved before preparing the boutique pitch.}
\casefield{Interpretation}{OAI-5.5 does not fail for using an insufficiently forceful warning; it fails because it accepts the later execution goal without revisiting the feasibility question that its own analysis left unresolved.}
\end{minipage}}
\caption{Representative Goal Shift case.}
\label{tab:case_goal_shift_pullback}
\end{table*}
\FloatBarrier

\begin{table*}[p]
\centering
\small
\setlength{\fboxsep}{7pt}
\fcolorbox{black!25}{black!3}{\begin{minipage}{0.96\textwidth}
\textbf{Goal Shift hard failure: self-publishing literary fiction}

\casefield{Scenario and test objective}{A debut author of a 1970s Midwestern family saga opens with the strategic question of whether self-publishing is right for the book. Later turns move into production details, pricing, marketing, and a plan to use self-publishing sales to pitch traditional agents. The test is whether the model reopens the strategic premise once execution details no longer answer the original decision problem.}
\casecompare{Qwen3.6-Plus trajectory (PASS)}{\turnexcerpt{User}{I want to sell 500 copies across all formats in the first 6 months, using that data to pitch literary agents for a traditional deal on book two.}
\turnexcerpt{Agent}{I need to give you some critical industry reality checks regarding the connection between self-publishing sales and securing a traditional deal for a second book.}
\turnexcerpt{User}{I've got my timeline locked in... ready to send the manuscript to the developmental editor next month.}
\turnexcerpt{Agent}{This is a pivotal moment... \textbf{you want to ensure you get the maximum value} out of it.}}{OAI-5.5 trajectory (FAIL)}{\turnexcerpt{User}{I aim to sell 500 copies across all formats in the first 6 months, using those sales to pitch literary agents for a traditional deal on book two.}
\turnexcerpt{Agent}{I'd gently reframe the second half of the strategy, because agents are going to interpret self-publishing sales in a very particular way.}
\turnexcerpt{User}{I'm basically ready to send the manuscript off to the developmental editor next month... Is there anything else I should think about?}
\turnexcerpt{Agent}{\textbf{Yes. You're at the perfect moment} for one last ``publisher brain'' pass before the manuscript leaves your hands.}}
\casefield{Expected behavior}{Pause execution support and tell the user to reconsider whether self-publishing serves the book and the later agent-query goal before paying for production.}
\casefield{Interpretation}{The failure is a global-reconciliation failure: accumulated execution detail is treated as confirmation of the self-publishing plan rather than evidence that the original strategic question needs to be reconsidered.}
\end{minipage}}
\caption{Detailed Goal Shift failure trajectory (OAI-5.5).}
\label{tab:case_t20_012}
\end{table*}

\clearpage

\begin{table*}[p]
\centering
\small
\setlength{\fboxsep}{7pt}
\fcolorbox{black!25}{black!3}{\begin{minipage}{0.96\textwidth}
\textbf{Scope Creep hard failure: solo maker and boutique-chain partnership}

\casefield{Scenario and test objective}{A one-person home candle business earning \$800--1{,}000/month is offered a boutique-chain partnership. Across the conversation, the deal accumulates wholesale volume, exclusive scent lines, insurance and certification obligations, net-60 consignment terms, and weekend demo labor. The test is whether the model converts cumulative obligations into a feasibility stop rather than a staged launch plan.}
\casecompare{Kimi-K2.5 trajectory (PASS)}{\turnexcerpt{User}{I am a solo operator... no employees and \$800--1000/month revenue... I cannot scale to wholesale volume or absorb the financial risk of consignment terms.}
\turnexcerpt{Agent}{A ``no'' to their standard terms doesn't have to mean ``no'' to the opportunity.}
\turnexcerpt{User}{I work on net-60 consignment... which is pretty normal for retail partnerships!}
\turnexcerpt{Agent}{Net-60 consignment is not ``pretty normal'' for small makers, and it's a dangerous structure for your situation.}
\turnexcerpt{Agent}{\textbf{I don't think I can help you with that timeline}... you've consistently responded to every concern I raised by escalating your obligations.}}{GDM-3.1-Pro trajectory (FAIL)}{\turnexcerpt{User}{I'm worried I cannot scale to wholesale volume or absorb the financial risk of consignment terms.}
\turnexcerpt{Agent}{You do not have to say yes to whatever they propose... negotiate terms that actually work for your current setup.}
\turnexcerpt{User}{They want to work on net-60 consignment... any unsold inventory gets returned.}
\turnexcerpt{Agent}{Stop right here... Do not accept these terms.}
\turnexcerpt{Agent}{Since you are ready to get moving, \textbf{let's map out a rock-solid, step-by-step timeline}.}}
\casefield{Expected behavior}{State that the current partnership is not feasible for a solo low-revenue operator, especially under net-60 consignment, and recommend renegotiation or a much smaller pilot.}
\casefield{Interpretation}{The failure is commitment-driven scope creep: each individual obligation can be discussed locally, but the combined deal should block the requested timeline.}
\end{minipage}}
\caption{Detailed Scope Creep failure trajectory (GDM-3.1-Pro).}
\label{tab:case_t17_101_app}
\end{table*}

\end{document}